\documentclass{article} 
\usepackage{iclr2027_conference,times}

\usepackage{amsmath,amsfonts,bm}

\def\eqref#1{equation~\ref{#1}}

\def\1{\bm{1}}

\DeclareMathAlphabet{\mathsfit}{\encodingdefault}{\sfdefault}{m}{sl}
\SetMathAlphabet{\mathsfit}{bold}{\encodingdefault}{\sfdefault}{bx}{n}

\usepackage{hyperref}
\usepackage{url}
\usepackage{tcolorbox}
\usepackage{xcolor}
\usepackage{colortbl} 
\usepackage{cleveref}
\usepackage{booktabs} 
\usepackage{multirow}
\usepackage{makecell} 
\usepackage{wrapfig}
\usepackage{amssymb}
\usepackage{times}             
\usepackage{inconsolata} 
\usepackage[T1]{fontenc}
\usepackage{inconsolata}
\usepackage{microtype}   
\usepackage{siunitx}
\usepackage{capt-of}
\usepackage{tabularx}
\usepackage{acticl_prompt_boxes}

\newsavebox{\tablepanelbox}
\newsavebox{\figurepanelbox}
\title{ActICL: Post-Training for Agentic Multimodal In-Context Learning}

\author{Yanshu Li$^{1}$\qquad
Ruixiang Tang$^{2}$\qquad
Junyi Jessy Li$^{1}$\qquad
Elias Stengel-Eskin$^{1}$\\
$^1$University of Texas at Austin \qquad
$^2$Rutgers University\\
}

\usepackage{xspace}

\newcommand{\method}[0]{ActICL\xspace}

\iclrfinalcopy 
\begin{document}

\maketitle

\begin{abstract}
Multimodal in-context learning (ICL) is an important capability of multimodal large language models (MLLMs). Its performance is closely tied to the demonstrations it uses, making demonstration retrieval a core problem in many ICL-based systems. Common methods use fixed scoring methods to retrieve all the demonstrations at once, which may introduce redundancy and noise. Context-aware retrieval methods have emerged to address this problem by allowing incremental retrieval based on a growing context. However, none of them lets the MLLM itself learn how to judge what new demonstrations it needs through its reasoning and then carry out progressive retrieval on its own. To this end, we introduce \textbf{ActICL}, a framework that trains an MLLM agent to acquire and use an ICL context via tool-augmented reasoning. ActICL builds an agent that analyzes its own information gaps during iterative reasoning and tool-use actions. Based on these gaps, it performs targeted retrieval to acquire a set of initial candidates. It can then use the existing context to further filter the candidates and refine the context, or it can iteratively perform a new retrieval. When the agent judges that the context is sufficient for the task, or when the resource limit is reached with the user request met, it sends the acquired context to an answering model, either the agent itself or another MLLM. We use a two-stage post-training procedure that jointly optimizes the agent's acquisition and answering. With an adaptive number of demonstrations, ActICL improves average performance by 11.6 percentage points over standard similarity-based retrieval across seven benchmarks and three MLLMs, and by 4.4 points compared to the strongest prior baseline. The contexts acquired by ActICL agents can also be used effectively by frontier models. When GPT-5.6-Sol performs 4-shot ICL, ActICL's contexts improve its performance by 3.5 points over standard retrieval, averaged over three benchmarks. Project page: 
\url{https://kaamava.github.io/acticl-project-page/}.
\end{abstract}

\section{Introduction}
\label{intro}
Multimodal in-context learning (ICL) enables multimodal large language models (MLLMs) to adapt to new tasks from demonstrations provided at inference time~\citep{Flamingo}, but its effectiveness depends heavily on whether these demonstrations form a context that helps the model solve the query~\citep{order, WMICLW}. Most retrieval-augmented ICL systems retrieve in a single step: they score each demonstration in an external bank independently, by embedding similarity~\citep{howto,CanMLLMs} or a probability-based score~\citep{selficl,revi}, and keep the top-ranked ones. Although efficient, this can yield noisy or otherwise suboptimal contexts~\citep{diveicl}. For instance, similarity-only retrieval can reduce diversity and induce shortcut learning driven by dominant features~\citep{howto}.

Thus, recent work has explored context-aware retrieval. Trained selectors estimate the utility of a demonstration conditioned on the current context~\citep{lever,grip,lsd}, but their implicit scoring functions are tied to their training distributions and generalize poorly to unseen tasks. Agentic workflows prompt a strong MLLM to adjust the context before passing it to an answering model~\citep{contextnav}. They bring reasoning into context construction but, as inference-time solutions, never learn from feedback what makes a context useful. Multimodal search agents do combine training with reasoning-based tool use~\citep{mmsearch,webwatcher}, but gather evidence that directly answers the query. Constructing ICL contexts instead requires assessing the instructional value of input--output demonstrations: whether each clarifies the query's task and what it adds beyond the current context. This motivates our central question: \textit{How can we train an MLLM to adaptively acquire useful context for agentic multimodal ICL, guided by its own reasoning and evolving information needs?}

\begin{figure}[t]
    \centering
    \includegraphics[width=0.93\textwidth]{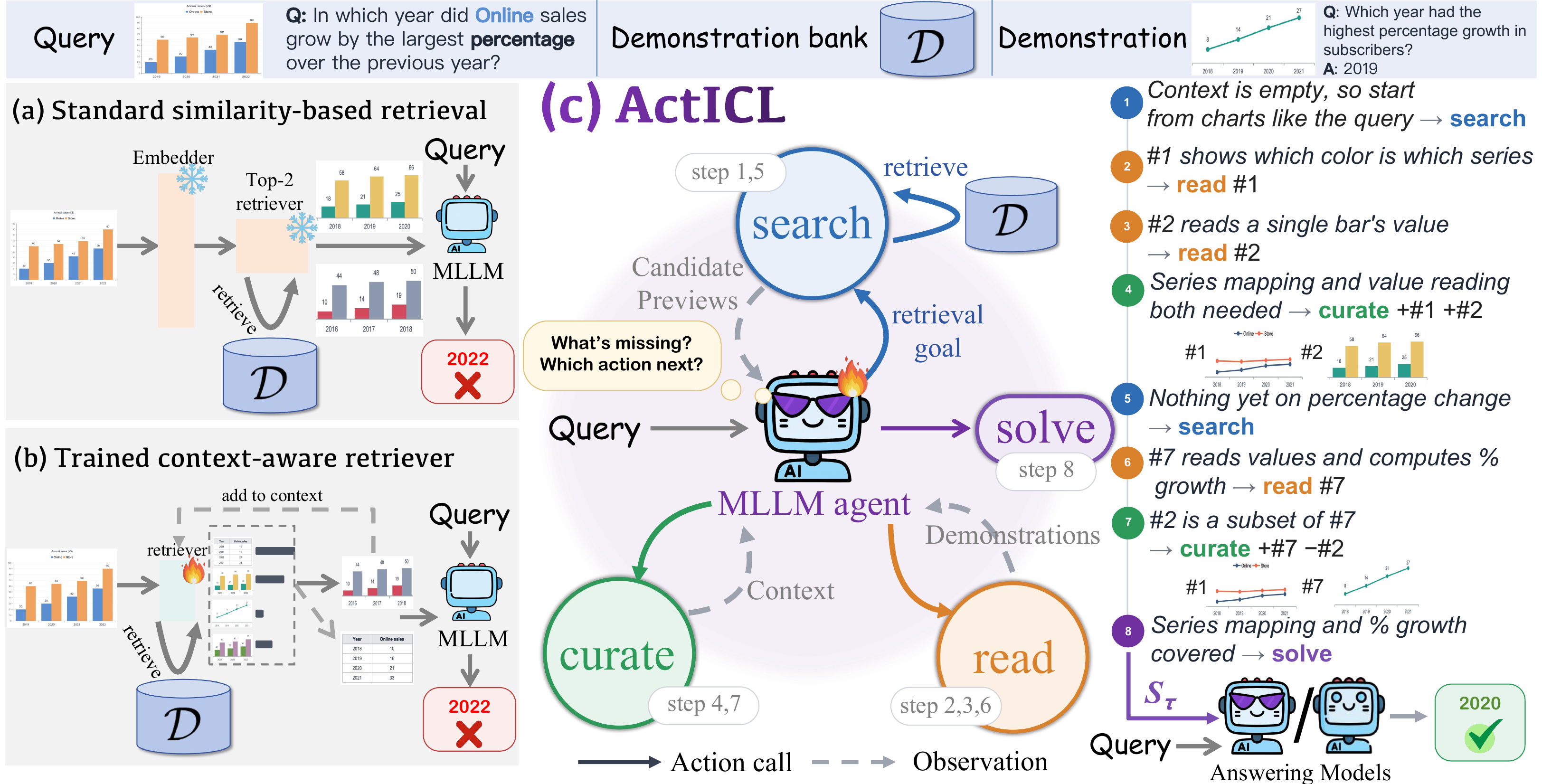}
    \caption{Given a multimodal query, we seek to construct a context with demonstrations that helps an MLLM answer the query correctly. 
    Prior work has largely (a) used off-the-shelf similarity-based retrievers, which often return redundant demonstrations, or (b) trained context-aware retrievers, which can fail to generalize at test time. 
    Our approach (c), \method{}, treats ICL context construction as an iterative agentic task, training an agent to formulate complex search queries, process and filter retrieved evidence, and decide when to submit a context to an answering model. 
    }
    \label{teaser}
    \vspace{-15pt}
\end{figure}

To address this question, we introduce \textbf{ActICL}, the first framework that trains an MLLM-based agent to construct and use a multimodal ICL context. 
As shown in Figure~\ref{teaser}, the agent interleaves reasoning with four tool-use actions (\texttt{search}, \texttt{read}, \texttt{curate}, and \texttt{solve}) to retrieve and inspect demonstrations, revise its context, and commit it.
During context acquisition, the agent decides whether to retrieve based on the test query and the demonstrations it has already inspected, obtaining new demonstration previews from the bank via \texttt{search}. Once enough candidates have been retrieved, the agent calls \texttt{read} to obtain the full content of key candidates. The \texttt{curate} action then lets it filter these demonstrations and update the context. 
Finally, when the agent adaptively considers the context sufficient, it invokes \texttt{solve} to commit it. 
To implement ActICL, we first design a pipeline that constructs training data covering diverse multimodal tasks with heterogeneous demonstration banks, exposing the agent to task-relevant demonstrations, distractors, and redundant information. On this dataset, we jointly optimize context acquisition and answering through a two-stage post-training procedure of rejection-sampling supervised fine-tuning (SFT) and reinforcement learning (RL), with rewards that combine answer quality and resource penalties.

We train ActICL agents with three different MLLMs as backbones: Qwen3-VL-8B, Qwen3-VL-30B-A3B, and Qwen3.5-9B, and evaluate on seven benchmarks.
Broadly, we find that ActICL improves performance across tasks and models. When the same ActICL-trained agent constructs the context and answers, \method{} improves accuracy
over standard similarity-based retrieval by 10.2 and 11.6 percentage points in the 4-shot and adaptive-shot settings, 
averaged across models and benchmarks. 
It also surpasses the strongest post-training-based baseline, ProMSA~\citep{promsa}, by 3.0 and 4.4 points. 
Moreover, we find that joint training also strengthens the model's internal ability to use multimodal context. With fixed ICL and interleaved image-text prompts, ActICL outperforms LoRA fine-tuning tailored to these tasks by 2.2 percentage points.
Finally, we find that ActICL agents can act as retrievers for stronger models. When constructing contexts for frontier models such as GPT-5.6-Luna and GPT-5.6-Sol, the ActICL agent built upon Qwen3.5-9B outperforms ContextNav~\citep{contextnav}, a test-time retrieval agentic workflow powered by Gemini3.5-Flash, by 2.1 points on average. In our ablation studies, we show that ActICL has a better marginal gain when adding more demonstrations, and that both stages of post-training are necessary. Together, these results show that reasoning-based training for context construction improves both context acquisition and utilization, with benefits that extend to broader ICL systems.

\section{Related Work}
\noindent\textbf{Multimodal agents.} 
Advances in the reasoning capabilities of multimodal large language models (MLLMs) have enabled them to select and invoke external tools at appropriate stages, giving rise to multimodal agents~\citep{llavaplus,pyvision, deepeyes}. By iteratively reasoning and calling tools, these agents can tackle increasingly complex tasks. Multimodal search agents exemplify this approach, sequentially interacting with external knowledge sources through policies trained with trajectory-level optimization~\citep{opensearch,webwatcher}. MMSearch-R1~\citep{mmsearch} introduces image and text retrieval tools and uses reinforcement learning to train agents for multi-turn search. DeepMMSearch-R1~\citep{deepmm} adds grounding and cropping tools to help agents focus on relevant content. Beyond web search, agents are emerging for entity-based visual question answering, such as ProMSA~\citep{promsa}. Our ActICL extends these post-training frameworks from gathering evidence to retrieving complete demonstrations.

\noindent\textbf{Multimodal in-context learning (ICL).} 
ICL~\citep{2020} is particularly valuable for MLLMs because it enables rapid test-time adaptation to new multimodal tasks from demonstrations without parameter updates~\citep{Flamingo,SurveyICL,many,rw1}. Motivated by analyses of multimodal ICL mechanisms~\citep{WMICLW, CanMLLMs}, researchers have explored ways to improve its robustness and performance. The primary focus of prior work has been on optimizing context construction~\citep{howto,efficl,mutualin}. These prior efforts design demonstration retrieval criteria but perform only a single static retrieval step without considering the model’s internal reasoning needs. Thus, recent work incorporates the answering model’s preferences and awareness of the current context into retrievers~\citep{lever,PMICL}. One of our key baselines, ContextNav~\citep{contextnav}, prompts a strong MLLM to orchestrate retrieval and refinement for a separate answering model, adjusting its workflow through inference-time feedback. Although ContextNav takes an important step toward agentic ICL, the judgment of context utility is not learned, and its reasoning filters candidates retrieved based on similarity rather than guiding the retrieval process as a whole. We instead train MLLM agents to internalize what makes context useful, such that they are capable of reasoning to determine what to retrieve, inspect, and retain.

\section{Method: ActICL}\label{sec:method}
\subsection{Overview}\label{sec:method-overview}
We study agentic multimodal ICL, in which the context is built iteratively by an MLLM agent alternating between reasoning and tool calls. 
Each instance has a query $q$, its ground-truth target $y^*$, and a demonstration bank $\mathcal{D}=\{(x_d,y_d)\}_{d=1}^{N_{\mathcal{D}}}$. Here, $q$ and each input $x_d$ contain one or more images and may also contain text, and $y_d$ is the target output for $x_d$. Given $q$, the agent $\pi_\theta$ iteratively acquires demonstrations from $\mathcal{D}$ through tool calls. 
The agent bases subsequent decisions on the already-acquired context. We call one full run of this process an acquisition. Its trajectory is denoted by $\tau$. The final demonstrations kept make up the context $S_\tau\subseteq\mathcal{D}$. 
In a new inference call, an answering model answers $q$ with the ICL instructions and $S_\tau$, producing a final answer $\hat{y}$. 
This answering model can be the same model backing the agent, or a different model.

\begin{figure}[t]
    \centering
    \includegraphics[width=\textwidth]{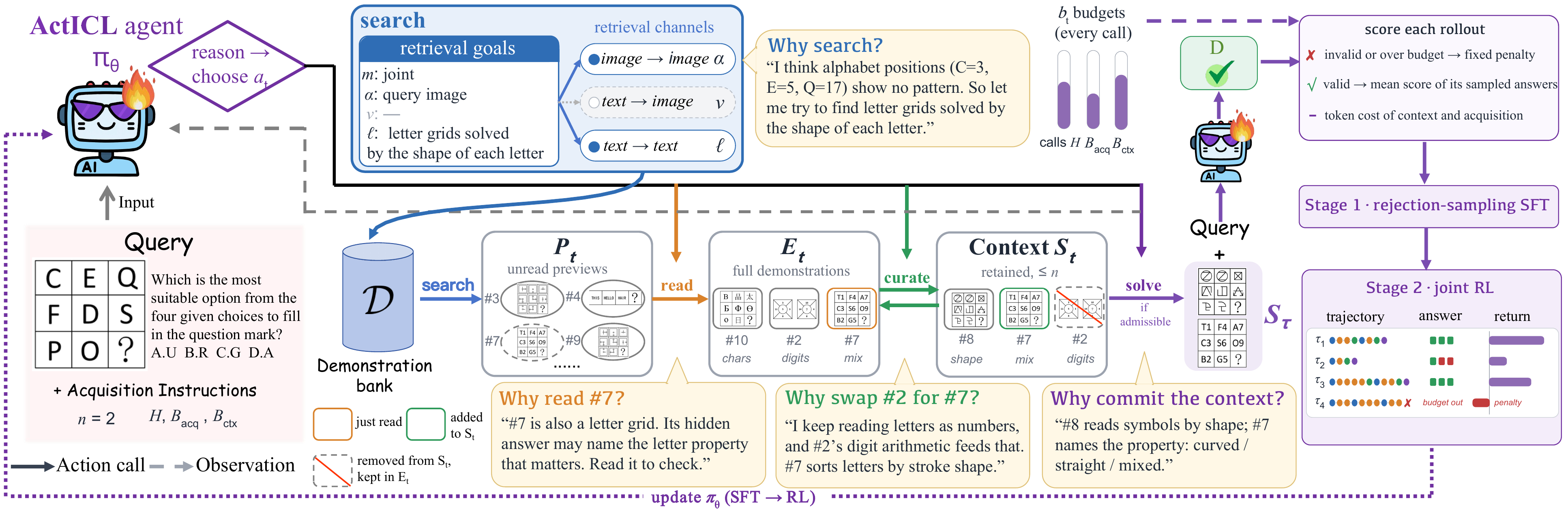} 
    \caption{
     The overall illustration of ActICL. Given a query and acquisition instructions, the agent reasons at every step to choose among four actions: \texttt{search} submits a goal whose enabled channels retrieve previews from the demonstration bank, \texttt{read} reveals full demonstrations, and \texttt{curate} adds or removes read demonstrations in the retained context, until \texttt{solve} commits an admissible context. The agent learns on-policy which contexts make ICL effective and how to acquire them.}
    \label{fig:main}
    \vspace{-10pt}
\end{figure}

We introduce \textbf{\method{}}, a framework for training MLLM agents in this setting, with its pipeline shown in Figure~\ref{fig:main}. The trained agent decides what to retrieve next from the bank, how to revise its context, and when to answer under resource constraints. It acquires demonstrations and revises the context with three actions, \texttt{search}, \texttt{read}, and \texttt{curate}, and invokes action \texttt{solve} to commit the context once it is ready to answer. The agent can then answer in a fresh call conditioned on the ICL instructions, $q$, and $S_\tau$, or pass $S_\tau$ to a different answering model. Training consists of supervised fine-tuning (SFT), followed by reinforcement learning (RL) to jointly optimize acquisition and answering.

\subsection{Context Acquisition Pipeline}\label{sec:method-context}
An \method{} agent builds its context step by step by interacting with a non-learned environment, which executes its tool calls and returns the results to the agent as observations. Users may specify a nonnegative integer $n$ to fix the number of demonstrations in the context, or leave it unspecified for adaptive acquisition. At each step $t$, the agent reasons over the preceding observation $o_{t-1}$ and issues a structured tool call $a_t$, which specifies an action type from
$\{\texttt{search},\texttt{read},\texttt{curate},\texttt{solve}\}$
together with its arguments, and then receives the next observation $o_t$. Thus, $\tau$ includes the initial input and the interleaved reasoning steps, actions, and observations up to commitment or termination. 

In agentic multimodal ICL, each demonstration contains images that are converted into many tokens, and these demonstrations keep accumulating during acquisition. To control the cost of acquisition, the user can set $H$ to bound the maximum number of actions in one acquisition. The user can also set $B_{\mathrm{acq}}$ as the total token budget of one acquisition and $B_{\mathrm{ctx}}$ as the token budget of the final context. All input and generated tokens of each acquisition call are counted against $B_{\mathrm{acq}}$.

\noindent\textbf{Search.}
When the agent requires more demonstrations, it first writes a retrieval goal $u_t$ and passes it to the \texttt{search} action. 
The environment then invokes a retrieval tool conditioned on $u_t$ to retrieve a set of candidate demonstrations from $\mathcal{D}$. We describe this retrieval procedure in detail in Section \ref{sec:method-retrieval}. 
To maintain a larger initial retrieval size while limiting the token cost, \texttt{search} returns only a token-limited preview of each candidate. 
Images are downsampled to thumbnails, and texts that exceed a length limit are truncated to their beginnings and ends. Appendix \ref{app:retrieval} gives the full rules for downsampling and truncation. We maintain a candidate pool $P_t$ that stores unread previews. $P_t$ is part of the environment and persists across searches, so the agent can return to earlier candidates. When a candidate first enters $P_t$, it is assigned a fixed integer index, allowing the agent to refer to each candidate in subsequent actions.

\noindent\textbf{Read.}
The agent calls \texttt{read} to expand retrieved candidates. Based on the query, the current context, and the demonstrations, it first selects the most task-relevant preview in $P_{t-1}$, which the environment expands into a full input--output pair and removes its preview from $P_t$.
We use a separate acquired set $E_t$ to store read demonstrations in their complete form from the bank.

\noindent\textbf{Curate.}
The agent uses \texttt{curate} to refine the context $S_{t-1}$ by reasoning over previous observations. 
\texttt{Curate} enables the agent to add demonstrations from $E_{t-1}$ to $S_{t-1}$ and to delete from $S_{t-1}$. The environment executes both add and delete together, letting the agent replace slots when the context is full. 
The removed demonstrations stay in $E_t$ and can be brought back without calling \texttt{read} again.

\noindent\textbf{Solve and context commitment.}
When the agent deems its current context sufficient, it invokes \texttt{solve} at step $t$. 
If $S_{t-1}$ is admissible, that is, it fits within $B_{\mathrm{ctx}}$ and, when $n$ is specified, 
contains exactly $n$ demonstrations, 
we use $S_{t-1}$ as the final context $S_\tau$. 
Otherwise, the \texttt{solve} call is rejected, returning an error message. 
The agent then continues acquiring context. 
When no further agent call is permitted, the environment commits the current context, if admissible. 
Otherwise, the acquisition ends in a budget failure, meaning the agent has run out of resources before reaching an admissible context. 
After $S_\tau$ is committed, an answering model is invoked. 
It can be the same \method{} agent, or another MLLM. 
The input of this call consists only of the user-provided ICL instructions, the query $q$, and $S_\tau$. 
The demonstrations in $S_\tau$ are in their complete form and ordered by when each was read.

\noindent\textbf{State and observation.}
After executing $a_t$, the environment updates the state and returns $o_t=(q,E_t,S_t,P_t,h_t,b_t)$,
where $h_t$ records reasoning, tool calls with their arguments, and structured feedback on execution status, affected or returned indices, and errors, and $b_t$ tracks the remaining calls and acquisition tokens, the remaining context capacity under $B_{\mathrm{ctx}}$, and, for fixed $n$, free context slots. A rejected call leaves $E_t$, $S_t$, and $P_t$ unchanged, but still counts toward $H$ and $B_{\mathrm{acq}}$. At each subsequent step, the environment constructs the agent input from $o_t$, as shown in Appendix~\ref{app:prompt}. ActICL's tool execution and resource accounting rules are detailed in Appendix~\ref{sec:tool-protocol}.

\subsection{Context-Aware Retrieval}\label{sec:method-retrieval}
A useful demonstration may resemble the query visually, or instead share its question type and solution procedure without image resemblance~\citep{ICLRetri}. Moreover, the relevant modality can shift as the context evolves. Thus, we provide multiple modality-specific retrieval channels, letting the agent target its current information needs. To support these channels, we precompute the image and text embeddings of all demonstration inputs in $\mathcal{D}$ with frozen encoders. At each \texttt{search} step, the agent reasons over $o_{t-1}$ to identify these needs and formulates a structured retrieval goal over $\mathcal{D}$: $u_t=(m_t,\alpha_t,v_t,\ell_t)$, where $m_t\in\{\mathrm{visual},\mathrm{textual},\mathrm{joint}\}$ specifies the retrieval mode, $\alpha_t$ denotes a set of reference images, and $v_t$ and $\ell_t$ are textual descriptions of the desired visual content and the demonstration task, respectively. For example, after acquiring demonstrations of chart-value reading, an agent answering a growth-related question may set $v_t$ to ``grouped bar charts with color-coded series across years'' and $\ell_t$ to ``questions comparing increase (\%) between years.''

The environment matches modalities to retrievers.
The $\mathrm{visual}$ mode has two retrieval channels. 
In the image-to-image channel, the agent selects one or more images from the query or acquired demonstrations in $E_{t-1}$
as the reference image set $\alpha_t$, with images scored based on similarity to reference images. 
In the text-to-image channel, the agent summarizes the needed visual information in natural language as $v_t$, and $v_t$ is then matched against the bank images.
The $\mathrm{textual}$ mode enables the text-to-text channel, allowing the agent to formulate queries flexibly through $\ell_t$, including the question type and the task type. $\mathrm{joint}$ mode enables all three channels. 
Within a mode, the agent may disable channels. 
Rankings from enabled channels are combined using a reciprocal-rank fusion rule.

\subsection{ActICL Training}\label{sec:method-training}
\subsubsection{Training Data}
We construct a set of 20k ICL instances. The task types in this set include knowledge and commonsense VQA (ReasonVQA, A-OKVQA, VCR~\citep{reasonvqa, aokvqa, vcr}), text-rich image understanding (DocVQA, ChartQA, TabMWP~\citep{docvqa,masry2022chartqa,lu2023tabmwp}), mathematical and diagram reasoning (GeoQA, ScienceQA, AI2D, CoSyn~\citep{geoqa, scienceqa,ai2d,cosyn}), and visual reasoning (BabyVision~\citep{babyvision}). We also use Omniglot, DTD, and CUB from Meta-Dataset~\citep{metad} for visual concept binding, and NLVR2 and Molmo2-SynMultiImageQA~\citep{nlvr2, molmo2} for native multi-image queries. When constructing the demonstration bank for each instance, we design a specific strategy so that the training data fully reflect the characteristics of ICL. We first remove exact duplicate cases, split each dataset into a query pool and a demonstration pool, and then merge all demonstration pools into a mixed pool of about 50k samples. Next, for each query, we sample from this mixed pool to construct a bank of size 512, 1024, or 4096. These sizes are chosen with probabilities of 20\%, 50\%, and 30\%, respectively. We sample without replacement through three routes applied in sequence: (i) sampling from the same task subset as the query within its source dataset, or from the dataset itself when no task subdivision is available; (ii) retrieving samples based solely on their fused visual and textual similarity to the query; and (iii) randomly sampling from datasets other than the query's source. We use these routes to construct a core of 256 demonstrations with varying route proportions, and fill the remaining bank positions with examples randomly sampled from all unselected examples in the pool. For visual concept binding, source-matched demonstrations are sampled from the instance's selected classes and use a consistent random label mapping. SFT and RL both draw from these instances. Appendix~\ref{app:training-data} details source proportions, bank sampling, and feasibility verification.

\subsubsection{Rejection-Sampling Supervised Fine-Tuning (SFT)}\label{sec:method-sft}
We first apply SFT to teach the agent to use tools in the prescribed format and to answer from the contexts it builds. Following outcome-filtered supervised training~\citep{deepseekai2025r1}, we hold each backbone's instruction-tuned checkpoint fixed and let it execute a fixed number of acquisition trajectories per instance. For every committed context, the same checkpoint then generates an answer. We keep trajectories with valid tool calls, compliant resource usage, and admissible final contexts, together with fully correct answers. Each backbone thus retains 4k trajectory--answer pairs from its own rollouts, as shown in Appendix~\ref{app:sft}. We jointly supervise acquisition and answering using the standard token-level cross-entropy (CE) loss over newly generated tokens. 
Each token is conditioned on its original call input and the preceding generated tokens within that call. 
Prompts, images, tool outputs, and replayed history serve only as conditioning inputs and are excluded from the loss.

\subsubsection{Joint Reinforcement Learning (RL)}
\label{sec:method-rl}
Starting from the SFT checkpoint, \method{} jointly optimizes acquisition and answering via RL with a clipped objective and group-relative baselines~\citep{shao2024deepseekmath}, mixing adaptive and fixed-count requests. 
We score each answer with the task's own evaluation metric, $r(\hat{y},y^*)\in[0,1]$. Acquisition is rewarded by how well its constructed context supports answering, minus token costs for the context and the acquisition process. 
These token-based costs encourage the agent to reach a compact yet effective context through as short an acquisition as possible, even at a fixed demonstration count. Acquisitions that fail to commit an admissible context receive a fixed negative return.

For each instance with its demonstration-count request and budgets, the current policy $\pi_{\theta_{\mathrm{old}}}$ samples $K$ acquisition trajectories $\tau_i$. Let $\delta_i\in\{0,1\}$ indicate whether $\tau_i$ is valid, i.e., commits an admissible context $S_i$ within the resource budgets. For each valid trajectory, the policy samples $M$ answers $\hat{y}_{ij}$, and we score $S_i$ by the mean task score $\bar r_i=M^{-1}\sum_{j=1}^{M} r(\hat{y}_{ij},y^*)$, reducing reliance on a single completion. The acquisition return is
\begin{equation}
    R_i=\begin{cases}
        \displaystyle \bar r_i
        -\lambda_{\mathrm{ctx}}\frac{T(S_i)}{B_{\mathrm{ctx}}}
        -\lambda_{\mathrm{acq}}
            \frac{\sum_t(I_{it}+O_{it})}{B_{\mathrm{acq}}},
        &\delta_i=1,\\
        -\kappa,&\delta_i=0,
    \end{cases}
    \label{eq:acquisition-return}
\end{equation}
where $T(S_i)$ is the token length of $S_i$, and $I_{it}$ and $O_{it}$ are the input and output tokens of the $t$-th acquisition call in $\tau_i$, with repeated observations counted each time they appear. Since budgets vary across instances, we divide each cost by its budget to keep the penalties on a common scale, then weight them by $\lambda_{\mathrm{ctx}},\lambda_{\mathrm{acq}}\ge 0$. We keep $\kappa>\lambda_{\mathrm{ctx}}+\lambda_{\mathrm{acq}}$, the highest cost a valid trajectory can incur, so that any valid trajectory outscores a failed one.

We compute acquisition and answer advantages separately. For acquisition, we compare each trajectory's return with those of the other $K-1$ trajectories for the same instance via a leave-one-out baseline~\citep{rloo}, giving $A_i^{\mathrm{acq}}$. For answering, we compare task rewards only among answers to the same context, so that differences reflect answer quality rather than shared acquisition costs. A purely group-relative baseline, however, yields zero advantage when all rewards in a group are identical, which is likely for tasks rewarding correct answers with $1$ and others with $0$ when few answers are sampled per context. Following fixed-reference advantages~\citep{nie2026gradient}, we therefore subtract a mixed baseline from each reward $r_{ij}=r(\hat{y}_{ij},y^*)$: $A_{ij}^{\mathrm{ans}}=r_{ij}-\big[(1-\mu)\,\bar r_{i,-j}+\mu\,b_{\mathrm{ref}}\big]$, where $\bar r_{i,-j}$ averages the other answers' rewards for $S_i$, $b_{\mathrm{ref}}=0.5$ is the midpoint of the reward range, and $0<\mu<1$. All-correct and all-incorrect groups then receive advantages of $\pm\mu/2$, while the remaining term preserves within-context comparisons. Let $\Phi_\theta(z,A)$ sum the clipped surrogate terms over the newly generated tokens of $z$ under advantage $A$. Because trajectory lengths vary widely, we normalize by a fixed constant $L_{\mathrm{ref}}>0$ instead of by length, avoiding inverse length weighting of token contributions. We jointly optimize
\begin{equation}
    \mathcal L_{\mathrm{RL}}(\theta) = -\frac{1}{L_{\mathrm{ref}}}\mathbb E_{\mathrm{batch}}\Bigg[\frac{1}{K}\sum_{i=1}^{K}\Phi_\theta(\tau_i,A_i^{\mathrm{acq}}) + \frac{1}{KM}\sum_{i:\,\delta_i=1}\sum_{j=1}^{M}\Phi_\theta(\hat{y}_{ij},A_{ij}^{\mathrm{ans}})\Bigg].
    \label{eq:joint-policy-loss}
\end{equation}
The two terms train acquisition and answering, respectively, with shared parameters $\theta$. Full rollout, advantage, clipping, and cost-schedule definitions are provided in Appendix~\ref{sec:rl-details}.

\section{Experiments}
\subsection{Setup}
We instantiate ActICL with Qwen3-VL-8B/30B-A3B-Instruct~\citep{qwenvl3} and Qwen3.5-9B. We use frozen SigLIP~\citep{zhai2023siglip} for image-to-image and text-to-image retrieval and Qwen3-Embedding-0.6B~\citep{Qwen3Emb} for text-to-text retrieval. We evaluate on seven benchmarks: TextVQA~\citep{singh2019textvqa}, MME-Reasoning (MME-R)~\citep{yuan2025mme}, CharXiv~\citep{wang2024charxiv}, GVL~\citep{GVL}, MathVision~\citep{wang2024mathvision}, MDK12~\citep{zhou2026mdk12}, and VL-ICL Bench~\citep{VL-ICL}. 
Together, they cover visual understanding, structured and quantitative reasoning, and context-dependent multimodal tasks while remaining disjoint from the training data.
When needed, we split demonstrations and queries 6:4 within each benchmark subcategory. 
At test time, each model's system prompt and each benchmark's ICL instruction remain fixed. 
The ICL instruction asks the answering model to output its answer directly. 
Our baselines include uniform random sampling from the bank, multimodal similarity-based MMICES~\citep{CanMLLMs}, ContextNav \citep{contextnav} with Gemini3.5-Flash as its policy, and LSD~\citep{lsd}, which trains retrievers with answering-model feedback. 
For LSD, we use the same data sources as ActICL and follow its original data construction method. 
We also adapt ProMSA~\citep{promsa}, a non-ICL multimodal retrieval agent trained with SFT and RL, for ICL context acquisition, using our data and a comparable training cost. 
Like ActICL, ProMSA uses its trained model to answer the query. We provide detailed hyperparameter settings and other implementation details in Appendices~\ref{app:sft}--\ref{app:evaluation}.

\subsection{Main Results}
\begin{table*}[t]
\caption{
Comparison of ICL performance with baseline methods. \textbf{Bold} indicates the best accuracy, and \underline{underlining} indicates the best accuracy excluding ActICL. The first row for each model reports zero-shot performance. All other results use four shots, except ActICL in the adaptive setting.
}
\label{tab:main1}
\centering
\resizebox{\linewidth}{!}{
\begin{tabular}{l*{8}{c}}
\toprule
\textbf{Methods}
& \textbf{TextVQA}
& \textbf{MME-R}
& \textbf{CharXiv}
& \textbf{GVL}
& \textbf{MathVision}
& \textbf{MDK12}
& \textbf{VL-ICL}
& \textbf{Avg.} \\
\midrule

Qwen3-VL-8B
& 0.735 & 0.238 & 0.473 & 0.554 & 0.528 & 0.462 & 0.293 & 0.469 \\
+Rand. Sample
& 0.728 & 0.259 & 0.482 & 0.527 & 0.540 & 0.449 & 0.562 & 0.507 \\
+MMICES
& 0.743 & 0.278 & 0.495 & 0.583 & 0.543 & 0.485 & 0.695 & 0.546 \\
+ContextNav
& 0.764 & 0.295 & 0.527 & 0.605 & 0.558 & 0.510 & 0.761 & 0.574 \\
+LSD
& 0.779 & 0.308 & 0.519 & 0.629 & 0.552 & 0.518 & 0.784 & 0.584 \\
+ProMSA
& \underline{0.819} & \underline{0.326} & \underline{0.546}
& \underline{0.684} & \underline{0.575} & \underline{0.561}
& \underline{0.873} & \underline{0.626} \\
\rowcolor{gray!15}
\textbf{+ActICL(4-shot)}
& 0.837 & 0.365 & 0.571 & 0.721 & 0.603 & 0.594
& \textbf{0.902} & 0.656 \\
\rowcolor{gray!15}
\textbf{+ActICL(adaptive)}
& \textbf{0.859} & \textbf{0.378} & \textbf{0.578}
& \textbf{0.739} & \textbf{0.616} & \textbf{0.620}
& 0.887 & \textbf{0.668} \\

\midrule

Qwen3-VL-30B-A3B
& 0.819 & 0.286 & 0.512 & 0.582 & 0.585 & 0.495 & 0.351 & 0.519 \\
+Rand. Sample
& 0.832 & 0.324 & 0.497 & 0.579 & 0.592 & 0.480 & 0.538 & 0.549 \\
+MMICES
& 0.851 & 0.308 & 0.508 & 0.594 & 0.597 & 0.492 & 0.746 & 0.585 \\
+ContextNav
& 0.863 & 0.335 & 0.536 & 0.616 & 0.618 & 0.521 & 0.825 & 0.616 \\
+LSD
& 0.865 & 0.352 & 0.529 & 0.624 & 0.642 & 0.529 & 0.803 & 0.621 \\
+ProMSA
& \underline{0.892} & \underline{0.388} & \underline{0.573}
& \underline{0.694} & \underline{0.660} & \underline{0.565}
& \underline{0.894} & \underline{0.667} \\
\rowcolor{gray!15}
\textbf{+ActICL(4-shot)}
& 0.922 & 0.430 & 0.602 & \textbf{0.739} & 0.693 & 0.597
& \textbf{0.910} & 0.699 \\
\rowcolor{gray!15}
\textbf{+ActICL(adaptive)}
& \textbf{0.928} & \textbf{0.445} & \textbf{0.623}
& 0.728 & \textbf{0.726} & \textbf{0.618}
& 0.904 & \textbf{0.710} \\

\midrule

Qwen3.5-9B
& 0.824 & 0.343 & 0.675 & 0.674 & 0.728 & 0.617 & 0.376 & 0.605 \\
+Rand. Sample
& 0.853 & 0.358 & 0.683 & 0.692 & 0.751 & 0.635 & 0.628 & 0.657 \\
+MMICES
& 0.859 & 0.366 & 0.680 & 0.705 & 0.759 & 0.649 & 0.814 & 0.690 \\
+ContextNav
& 0.870 & 0.387 & 0.702
& \underline{0.732} & 0.781 & 0.673
& 0.877 & 0.717 \\
+LSD
& 0.868 & 0.396 & 0.697 & 0.719
& 0.785 & 0.675 & 0.845 & 0.712 \\
+ProMSA
& \underline{0.908} & \underline{0.405} & \underline{0.736}
& 0.728 & \underline{0.829} & \underline{0.714}
& \underline{0.889} & \underline{0.744} \\
\rowcolor{gray!15}
\textbf{+ActICL(4-shot)}
& \textbf{0.935} & 0.427 & 0.751 & 0.778 & 0.862 & 0.737
& 0.913 & 0.772 \\
\rowcolor{gray!15}
\textbf{+ActICL(adaptive)}
& 0.929 & \textbf{0.468} & \textbf{0.788}
& \textbf{0.794} & \textbf{0.870} & \textbf{0.760}
& \textbf{0.921} & \textbf{0.790}\\

\bottomrule
\end{tabular}
}
\vspace{-0.12in}
\end{table*}

\textbf{ActICL consistently improves multimodal ICL with both fixed-shot and adaptive context acquisition.}
We first compare context acquisition strategies on the three backbones used to train ActICL agents. 
As shown in Table~\ref{tab:main1}, 4-shot ActICL outperforms all baselines across all seven benchmarks and three backbones, with average gains of 8.2--11.4 percentage points over MMICES and 2.8--3.2 points over ProMSA. Adaptive acquisition further improves average accuracy by 1.1--1.8 percentage points while using an average of 3.54 shots. These results demonstrate the effectiveness of ActICL at a matched demonstration count, and the additional benefits of teaching the model to perform agentic ICL effectively and to adapt context construction to its reasoning needs.
\begin{table*}[t]
\caption{Performance comparison with baseline methods on three multimodal ICL tasks and two multi-image understanding tasks. All few-shot prompts are constructed using MMICES.}
\label{tab:main2}
\centering
\small
\setlength{\tabcolsep}{3pt}       
\renewcommand{\arraystretch}{0.9}
\setlength{\aboverulesep}{0.3ex} 
\setlength{\belowrulesep}{0.3ex} 
\begin{tabular}{l*{8}{c}}
\toprule
\multirow{2}{*}{\textbf{Method}}
& \multicolumn{2}{c}{\textbf{MME-R}}
& \multicolumn{2}{c}{\textbf{MathVision}}
& \multicolumn{2}{c}{\textbf{MMStar}}
& \multirow{2}{*}{\textbf{MuirBench}}
& \multirow{2}{*}{\textbf{BLINK}} \\
\cmidrule(lr){2-3}
\cmidrule(lr){4-5}
\cmidrule(lr){6-7}
& 4-shot & 8-shot
& 4-shot & 8-shot
& 4-shot & 8-shot
& & \\
\midrule

Qwen3-VL-8B
& 0.278& 0.316&0.543 & 0.557&0.746& 0.753& 0.653&0.702 \\
+LoRA
& 0.295& 0.324& 0.536& 0.569& 0.762& 0.784&0.709 &0.735 \\
+DARA
& 0.306& 0.305&0.529 &0.568 &0.770 & 0.769& 0.684&0.719\\
\rowcolor{gray!15}
\textbf{+ActICL}
& \textbf{0.344}&\textbf{0.375}& \textbf{0.591}& \textbf{0.584}& \textbf{0.783}& \textbf{0.797}& \textbf{0.717}& \textbf{0.744}\\

\midrule

Qwen3-VL-30B-A3B
& 0.308&0.332& 0.597& 0.609&0.751&0.746 & 0.629& 0.683\\
+LoRA
& 0.346& 0.358& 0.613&0.632& 0.785& 0.792&0.647 & 0.695\\
+DARA
& 0.352& 0.365& 0.625& 0.627&0.796 & 0.788& 0.608&0.688\\
\rowcolor{gray!15}
\textbf{+ActICL}
&\textbf{0.365} & \textbf{0.380}&\textbf{0.653} &\textbf{0.648} & \textbf{0.807}&\textbf{0.809}& \textbf{0.661}& \textbf{0.708}\\

\midrule

Qwen3.5-9B
& 0.366&0.351 & 0.759&0.762 & 0.826& 0.833& 0.738& 0.816\\
+LoRA
& 0.375& 0.363& 0.762& 0.776&0.835&0.829& 0.754& 0.837\\
+DARA
& 0.382& 0.359& 0.770& 0.781& 0.841& 0.837& 0.719& 0.820\\
\rowcolor{gray!15}
\textbf{+ActICL}
& \textbf{0.397}& \textbf{0.373}&\textbf{0.785} & \textbf{0.790}&\textbf{0.860} & \textbf{0.843}& \textbf{0.767}&\textbf{0.851} \\

\bottomrule
\end{tabular}
\vspace{-0.1in}
\end{table*}

\textbf{ActICL also strengthens the model's reasoning over interleaved image--text contexts.}
We evaluate answering without ActICL's acquisition loop. To construct complex multimodal contexts, we use MMICES to build 4- and 8-shot prompts on MME-R, MathVision, and MMStar~\citep{MMStar}. We add two more benchmarks designed for interleaved image-text tasks, MuirBench~\citep{muirbench} and BLINK~\citep{blink}. For each of the three backbones used in the main experiments, we compare the ActICL-trained model with two baselines built on the same backbone, which are fine-tuned with LoRA~\citep{lora} and DARA~\citep{dara}, respectively, to improve interleaved image-text understanding. We control their training data sources and sizes to ensure a fair comparison, as detailed in Appendix~\ref{app:evaluation}. As shown in Table~\ref{tab:main2}, ActICL achieves the best results across all backbones and settings. Its largest gain over the strongest competing baseline is 5.1 percentage points on 8-shot MME-R with Qwen3-VL-8B. Because these contexts are supplied externally and remain fixed, the improvements indicate stronger reasoning over complex context. They support the broader benefits of the training and extend ActICL's utility to more real-world tasks.

\textbf{ActICL remains effective when used solely to construct contexts for other models.} To evaluate acquisition independently of the trained agent’s answering ability, we use ActICL-trained Qwen3.5-9B to construct 4-shot contexts for Qwen3.5-397B-A17B, GPT-5.6-Luna, and GPT-5.6-Sol on three challenging visual understanding benchmarks for frontier MLLMs: 
\begin{wraptable}{r}{0.55\linewidth}
\centering
\caption{4-shot performance of different retrievers with larger-scale open-source and frontier closed-source models. ContextNav uses Gemini3.5-Flash as its policy, while ActICL uses Qwen3.5-9B.}
\label{tab:main3}
\footnotesize
\setlength{\tabcolsep}{4pt}
\renewcommand{\arraystretch}{1.05}
\resizebox{\linewidth}{!}{
\begin{tabular}{lccc}
\toprule
\textbf{Method}
& \textbf{PerceptionBench}
& \textbf{MindCube}
& \textbf{MMMU-Pro} \\
\midrule

Qwen3.5-397B-A17B & 0.458& 0.477& 0.757\\
+MMICES     & 0.472& 0.492&0.761\\
+ContextNav          & 0.487& 0.524& 0.772\\
\rowcolor{gray!15}
\textbf{+ActICL} &\textbf{0.500}& \textbf{0.545}&\textbf{0.780} \\

\midrule

GPT-5.6-Luna (max) & 0.559& 0.638& 0.785\\
+MMICES    & 0.568&0.634 & 0.802\\
+ContextNav    & 0.590& 0.669& 0.809\\
\rowcolor{gray!15}
\textbf{+ActICL} & \textbf{0.615}& \textbf{0.697}& \textbf{0.820}\\

\midrule

GPT-5.6-Sol (max)  & 0.624&0.716 & 0.836\\
+MMICES   &0.636 &0.728 & 0.844\\
+ContextNav   &0.643&0.740 &0.849 \\
\rowcolor{gray!15}
\textbf{+ActICL} & \textbf{0.674}&\textbf{0.765} & \textbf{0.874}\\

\bottomrule
\end{tabular}
}
\vspace{-3em}
\end{wraptable}
PerceptionBench~\citep{perceptionbench}, MindCube~\citep{mindcube}, and MMMU-Pro~\citep{mmmupro}. As shown in Table~\ref{tab:main3}, ActICL leads across all nine model--benchmark combinations, exceeding ContextNav by 0.8--3.1 percentage points, with the largest gain observed for GPT-5.6-Sol. These gains with fixed answering models establish a contribution from context construction itself and support the effectiveness of reasoning-guided acquisition, beyond improvements to the agent's own answering ability. They also show the value of a relatively small ActICL agent as a context constructor for commercial-scale models, which allows it to be integrated into diverse and evolving multimodal systems.
\subsection{Ablation Studies}
\label{sec:ablations}
\begin{figure*}[t]
\centering
\begin{minipage}[c]{0.48\textwidth}
\centering
\captionof{table}{Ablation on training stages with Qwen3.5-9B under the adaptive setting. ``w/o acquisition'' and ``w/o answering'' run joint RL after SFT without the corresponding part.}
\label{tab:ablation-stages}
\vspace{6pt}
\footnotesize
\setlength{\tabcolsep}{4pt}
\renewcommand{\arraystretch}{1.05}
\resizebox{\linewidth}{!}{%
\begin{tabular}{@{}lcccc@{}}
\toprule
\textbf{Method}
& \textbf{MME-R}
& \textbf{CharXiv}
& \textbf{MathVision}
& \textbf{Avg.} \\
\midrule
Base
& 0.343 & 0.675 & 0.728 & 0.582 \\
SFT only
& 0.396 & 0.740 & 0.835 & 0.657 \\
Joint RL only
& 0.379 & 0.734 & 0.794 & 0.636 \\
\midrule
w/o acquisition
& 0.451 & 0.762 & 0.853 & 0.689 \\
w/o answering
& 0.437 & 0.769 & 0.845 & 0.684 \\
\method{}
& \textbf{0.468}
& \textbf{0.788}
& \textbf{0.870}
& \textbf{0.709} \\
\bottomrule
\end{tabular}%
}
\par
\end{minipage}\hfill
\begin{minipage}[c]{0.48\textwidth}
\centering
\includegraphics[width=\linewidth]{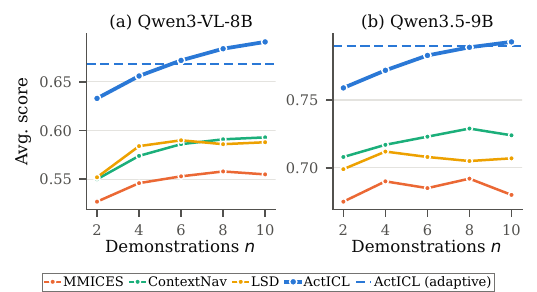}
\par
\captionof{figure}{Average score versus the number of demonstrations. The dashed line marks \method{} with adaptive acquisition.}
\label{fig:ablation-shots}
\end{minipage}
\vspace{-0.5em}
\end{figure*}

\noindent\textbf{Training stages.}
We ablate the training stages of Qwen3.5-9B under the adaptive setting on three benchmarks. Besides applying SFT or joint RL alone, we run joint RL after SFT without acquisition or without answering rewards, which removes the corresponding term of Eq.~\ref{eq:joint-policy-loss}.
As shown in Table~\ref{tab:ablation-stages}, each training stage helps on its own. SFT raises the average score from 0.582 to 0.657, and joint RL alone reaches 0.636. Joint RL from the base checkpoint still trails SFT by 2.1 percentage points, suggesting that the policy benefits from learning the tool protocol before optimizing outcomes. Starting from SFT, joint RL without acquisition or without answering improves the average, but both remain below the full ActICL. Removing answering lowers the average by 2.5 points and removing acquisition by 2.0 points, so the policy must learn both which demonstrations to retain and how to exploit them. The full ActICL is best on every benchmark.
Appendix~\ref{app:ablations} reports the 4-shot setting.

\noindent\textbf{Demonstration count in the constructed context.} Adding demonstrations often brings diminishing returns, as static retrieval increasingly adds redundant or distracting examples. We compare \method{} with MMICES, ContextNav, and LSD at $n\in\{2,4,6,8,10\}$ in Figure~\ref{fig:ablation-shots}, where $n=6$ and $n=10$ are unseen in ActICL training. The baselines saturate early, gaining at most 1.9 percentage points beyond four demonstrations, and MMICES and LSD even decline on Qwen3.5-9B. In contrast, \method{} improves at every step. Its own per-step gains shrink but stay positive through the unseen counts. Adaptive acquisition commits only 3.93 and 3.26 demonstrations on average, yet it surpasses fixed 4-shot contexts on both models and matches eight shots on Qwen3.5-9B.

\begin{wrapfigure}{r}{0.46\textwidth}
\vspace{-20pt}
\centering
\includegraphics[width=\linewidth]{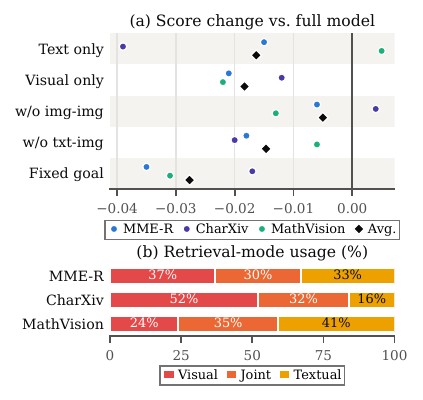}
\vspace{-16pt}
\caption{Multimodal retrieval ablation on Qwen3.5-9B under the adaptive setting.}
\label{fig:ablation-retrieval}
\vspace{-8pt}
\end{wrapfigure}
\textbf{Context-aware multimodal retrieval.} We disable each retrieval channel at test time and, separately, replace the agent's own goals with a fixed query-based goal.
As Figure~\ref{fig:ablation-retrieval}(a) shows, Fixed goal causes the largest drop, lowering the average by 2.8 percentage points and MME-R by 3.5 points, where the agent normally uses the three modes almost evenly (Figure~\ref{fig:ablation-retrieval}(b)).
The diversity of sources needed in this setting makes the ability to reformulate goals as the context grows especially valuable.
Keeping a single modality lowers the average by 1.6 points (Text only) and 1.8 points (Visual only).
Text only loses 3.9 points on CharXiv, where 52\% of searches are visual, whereas Visual only loses 2.2 points on MathVision, where textual retrieval is most frequent.
Within the visual channels, removing text-to-image retrieval costs more than removing image-to-image retrieval, with the largest loss of 2.0 points on CharXiv.
Describing queries in language lets the agent request new charts, whereas image-to-image retrieval only finds inputs resembling images it already has. Appendix \ref{app:ablations} presents additional ablations on ActICL’s \texttt{read} and \texttt{curate} actions, reward design, and training instance construction, and a failure mode study.

\section{Conclusion}
We presented \method{}, a framework that post-trains MLLM-based agents to acquire and use multimodal ICL contexts through reasoning and tool use. ActICL training improves both demonstration acquisition and multi-image understanding with provided contexts. It strengthens the ability of MLLMs to obtain and use multimodal contextual information on their own. The trained agents also construct contexts for frontier models, extending these benefits to closed-source systems. These findings highlight ActICL training as a path toward more capable autonomous multimodal agents.

\section*{Acknowledgments}
The authors acknowledge the Texas Advanced
Computing Center (TACC) at The University of Texas at Austin for providing computational resources that have contributed to the research results reported within this paper.
\bibliography{iclr2027_conference}
\bibliographystyle{iclr2027_conference}

\appendix
\section{Environment and Tool Protocol}\label{sec:tool-protocol}
This appendix formalizes the environment of Sections~\ref{sec:method-context} and~\ref{sec:method-retrieval}. The agent may interleave \texttt{search}, \texttt{read}, and \texttt{curate} in any order while resources remain.
 
\subsection{Retrieval and Candidate Previews}\label{app:retrieval}
 
\noindent\textbf{Index and encoders.}
The backend indexes only the input $x_d$ of each record, so the agent must \texttt{read} a candidate before it can see the target $y_d$. We write $\mathcal G_d$ for the ordered images of $x_d$ and $w_d$ for its concatenated input text. Three frozen encoders embed these inputs. SigLIP (siglip-so400m-patch14-384) provides the image encoder $f_I$ and the aligned text encoder $f_V$, and Qwen3-Embedding-0.6B provides the text-retrieval encoder $f_T$. The backend $\ell_2$-normalizes all embeddings, precomputes them once for every record of the mixed demonstration pool (Appendix~\ref{app:training-data}) and of each evaluation bank, and runs exact similarity search over every bank. Freezing the encoders keeps the retrieval environment fixed while the policy changes during training.
 
\noindent\textbf{Retrieval channels.}
A \texttt{search} call carries the goal $u_t=(m_t,\alpha_t,v_t,\ell_t)$ of Section~\ref{sec:method-retrieval}. For each record $d\in\mathcal D$, the three channels compute
\begin{equation}
    \begin{aligned}
        s_{\alpha,t}(d)&=\max_{g\in\mathcal G_d}\,
            \frac{1}{|\alpha_t|}\sum_{e\in\alpha_t}
            \langle f_I(e),f_I(g)\rangle,\\
        s_{v,t}(d)&=\max_{g\in\mathcal G_d}\,
            \langle f_V(v_t),f_I(g)\rangle,\\
        s_{\ell,t}(d)&=\langle f_T(\ell_t),f_T(w_d)\rangle.
    \end{aligned}
    \label{eq:retrieval-channels}
\end{equation}
The image-to-image channel first averages the similarity of a bank image to all reference images and then scores the record by its best image. The text-to-image channel also takes the best image. This maximum prevents the extra images of a multi-image input from diluting a relevant match. Every input contains at least one image, so the two image channels consider all records, $\mathcal D_\alpha=\mathcal D_v=\mathcal D$. The text-to-text channel considers only the subset $\mathcal D_\ell\subseteq\mathcal D$ of records with nonempty input text.
 
\noindent\textbf{Channel activation.}
The mode determines which channels a search may use, and the agent can switch off a channel by leaving its goal field empty when that field would not express its current need. Let $\Gamma(u_t)\subseteq\{\alpha,v,\ell\}$ contain the channels whose goal fields are nonempty. The environment activates
\begin{equation}
    \begin{gathered}
    \Omega_t=\mathcal M(m_t)\cap\Gamma(u_t),\\
    \mathcal M(\mathrm{visual})=\{\alpha,v\},\quad
    \mathcal M(\mathrm{textual})=\{\ell\},\quad
    \mathcal M(\mathrm{joint})=\{\alpha,v,\ell\},
    \end{gathered}
    \label{eq:channel-activation}
\end{equation}
and rejects the search when $\Omega_t=\varnothing$. The acquisition instructions ask the agent to leave fields outside the selected mode empty (Appendix~\ref{app:prompt}), and the intersection in Eq.~\ref{eq:channel-activation} ignores any that are not. For each $\omega\in\Omega_t$, the backend forms the list $L_{\omega,t}$ of the $N_{\mathrm{list}}=25$ records in $\mathcal D_\omega$ with the highest $s_{\omega,t}$ and breaks ties by the bank's fixed insertion order for reproducibility.
 
\noindent\textbf{Rank fusion and candidate selection.}
Similarities from different encoders lie on different scales, so the backend fuses ranks instead of calibrating scores. It applies reciprocal-rank fusion,
\begin{equation}
    s_t(d)=\sum_{\omega\in\Omega_t:\,d\in L_{\omega,t}}
        \frac{1}{\beta+\operatorname{rank}_{\omega,t}(d)},
    \label{eq:retrieval-fusion}
\end{equation}
where $\operatorname{rank}_{\omega,t}(d)\in\{1,\dots,N_{\mathrm{list}}\}$ is the position of $d$ in $L_{\omega,t}$ and $\beta=60$. Let $\mathcal U_{t-1}=\{d\in\mathcal D:(x_d,y_d)\in E_{t-1}\text{ or }\tilde x_d\in P_{t-1}\}$ contain the records the agent has already encountered, where $\tilde x_d$ denotes the preview defined below. The backend returns
\begin{equation}
    \mathcal C_t=\operatorname{Top}_{N_{\mathrm{cand}}}\Big(
        \big(\textstyle\bigcup_{\omega\in\Omega_t}L_{\omega,t}\big)\setminus\mathcal U_{t-1};\ s_t\Big),
    \qquad N_{\mathrm{cand}}=6,
    \label{eq:retrieval-page}
\end{equation}
where $\operatorname{Top}_N(X;s)$ lists the $N$ elements of $X$ with the highest $s$ under the same tie-breaking rule, or all of $X$ when $|X|<N$. Excluding $\mathcal U_{t-1}$ spends every search on unseen records, and the agent can still read earlier candidates from $P_{t-1}$. The cap $N_{\mathrm{cand}}$ bounds how fast previews accumulate in later observations while still offering several alternatives per search. Because each channel list stops at $N_{\mathrm{list}}$, repeating a goal eventually exhausts its candidates, and the agent must reformulate the goal to reach other records. A search that finds no unseen record returns $\mathcal C_t=\varnothing$ and still counts as an executed call.
 
\noindent\textbf{Preview construction.}
For each $d\in\mathcal C_t$, the environment builds an input-only preview $\tilde x_d$ without another model call. A preview keeps at most two images within 64 visual tokens and at most 128 tokens of input text, which exposes enough structure for the agent to decide what to read at a small and predictable cost. 
 
For a multi-image input, the environment keeps the first and last images in their original order and splits the visual allowance equally between them, so that both ends of the input remain visible. A single-image input uses the full allowance. The environment resizes each kept image with the backbone's processor, preserving its aspect ratio up to the processor's dimension rounding, and verifies the resulting token count. It never crops, so each thumbnail retains the global layout of its image. The preview also reports the total image count and the kept and omitted positions, which makes every omission explicit to the agent.
 
The environment concatenates the input-text blocks in their original order and tokenizes them with the backbone's tokenizer. Text within 128 tokens appears in full. For longer text, the environment reserves space for the marker \texttt{[Text omitted]} and splits the remaining allowance between a prefix and a suffix in a $3{:}1$ ratio, rounding the prefix length down. The prefix preserves opening task instructions, and the suffix preserves trailing questions or answer options. A text-free input yields an empty excerpt. Because the construction is deterministic, a candidate shows the same preview in every observation until the agent reads it. 
 
\subsection{Tool Execution and Observations}\label{app:tool-execution}
 
\noindent\textbf{Candidate indices and image references.}
The environment numbers candidates from one in order of first exposure and binds each number to its record for the whole acquisition. Stable numbers let the agent refer to the same candidate in \texttt{read} and \texttt{curate} without tracking any renumbering. To place an image in $\alpha_t$, the agent names either a position in $q$ or an acquired demonstration's number together with an image position within it. The environment accepts image references only to $q$ and $E_{t-1}$, because the agent has not yet inspected unread candidates in full.
 
\noindent\textbf{State transitions.}
An accepted \texttt{search} adds the new previews and leaves the rest of the state unchanged:
\begin{equation}
    P_t=P_{t-1}\cup\{\tilde x_d:d\in\mathcal C_t\},\qquad E_t=E_{t-1},\qquad S_t=S_{t-1}.
    \label{eq:appendix-search-transition}
\end{equation}
An accepted \texttt{read} of a candidate with $\tilde x_d\in P_{t-1}$ moves that candidate from the preview pool to the acquired set:
\begin{equation}
    E_t=E_{t-1}\cup\{(x_d,y_d)\},\qquad P_t=P_{t-1}\setminus\{\tilde x_d\},\qquad S_t=S_{t-1}.
    \label{eq:appendix-read-transition}
\end{equation}
A \texttt{curate} call names removals $S_t^-\subseteq S_{t-1}$ and additions $S_t^+\subseteq E_{t-1}\setminus S_{t-1}$. The environment forms the candidate context
\begin{equation}
    S'_t=(S_{t-1}\setminus S_t^-)\cup S_t^+
    \label{eq:appendix-curate-transition}
\end{equation}
and sets $S_t=S'_t$ if $T(S'_t)\le B_{\mathrm{ctx}}$ and, under a fixed count, $|S'_t|\le n$. It keeps $E_t=E_{t-1}$ and $P_t=P_{t-1}$. A demonstration that the agent chose to read remains relatively important after it leaves the context, for example as a fallback when a replacement proves weaker, so it stays available for later reconsideration without another \texttt{read}. Because additions come only from $E_{t-1}$ and the acquired set never shrinks, $S_t\subseteq E_t$ holds at every step.
 
\noindent\textbf{Invalid calls and history.}
Besides malformed calls, the environment rejects unavailable indices or image references, \texttt{curate} updates that violate the constraints above, and inadmissible \texttt{solve} calls. A generation that exhausts the acquisition allowance $G_{\mathrm{acq}}$ before completing a tool call also counts as malformed. The environment records every error in the history, which stores only reasoning, calls, arguments, and feedback. Demonstration and preview contents reach the agent through $E_t$ and $P_t$ alone, which keeps a single copy of each in the observation. 
 
\noindent\textbf{Observation serialization.}
The user-defined inputs of an acquisition call are the acquisition instructions, which contain the acquisition protocol and the tool definitions, and the count request. The environment places the serialized observation $o_{t-1}$ between them. It presents the query first. It then lists the acquired demonstrations once in first-read order and marks those in $S_{t-1}$ instead of repeating the context. Unread previews follow in first-exposure order, then the history and the remaining resources $b_{t-1}$, and the count request closes the input. The environment rebuilds this input in full at every call and never truncates or summarizes demonstrations, previews, or history, so the agent can revisit any earlier judgment. Appendix~\ref{sec:acquisition-protocol} specifies what happens when the input no longer fits.
 
\subsection{Resource Accounting and Commitment}\label{sec:acquisition-protocol}
 
\noindent\textbf{Token accounting.}
The environment counts text and visual tokens after serialization. We write $T(S)$ for the token length of the demonstrations in $S$ including their presentation delimiters, and $I_t$ and $O_t$ for the input and generated tokens of acquisition call $t$. After call $t$, the environment reports
\begin{equation}
    b_t=\Big(H-t,\;
        B_{\mathrm{acq}}-\textstyle\sum_{s=1}^{t}(I_s+O_s),\;
        B_{\mathrm{ctx}}-T(S_t),\;
        n-|S_t|\Big),
    \label{eq:resource-state}
\end{equation}
starting from $b_0=(H,B_{\mathrm{acq}},B_{\mathrm{ctx}},n)$ and dropping the last component in adaptive mode. The environment charges demonstrations, previews, and history each time they enter an input, because keeping information visible to the agent costs tokens at every call. It deducts the actual generated length $O_t$.
 
\noindent\textbf{Call admission.}
Before issuing call $t$, the environment computes the complete input length $I_t$, which covers the acquisition instructions, the query, acquired demonstrations, previews, history, and resource metadata. It admits the call only if
\begin{equation}
    t\le H,\qquad
    I_t+G_{\mathrm{acq}}\le W,\qquad
    \sum_{s<t}(I_s+O_s)+I_t+G_{\mathrm{acq}}\le B_{\mathrm{acq}},
    \label{eq:appendix-path-check}
\end{equation}
where $W$ is the per-call window and $G_{\mathrm{acq}}$ is the generation allowance of an acquisition call. Reserving the full allowance guarantees that every admitted call can complete its response within both limits. Since $O_t\le G_{\mathrm{acq}}$, every admitted call also satisfies $\sum_{s\le t}(I_s+O_s)\le B_{\mathrm{acq}}$, so the cumulative acquisition cost never exceeds its budget. When the next call fails admission, acquisition ends under the commitment rule of Section~\ref{sec:method-context}, since truncation would silently change what the agent knows. 
 
\noindent\textbf{Windows and image processing.}\label{app:resource-settings}
All backbones share the per-call window $W=65{,}536$ tokens, including visual tokens, so that resource comparisons use the same allowance. Each acquisition call may generate $G_{\mathrm{acq}}=1{,}024$ tokens and each answering call $G_{\mathrm{ans}}=8{,}192$ tokens, both including reasoning. The environment counts actual visual tokens under the native 16-pixel patches with $2\times2$ spatial merging. Query images share 4,096 visual tokens, raised to 8,192 for dense multi-image queries. Each complete demonstration receives 2,048 visual tokens across its images, raised to 4,096 for dense documents, charts, BabyVision, and multi-image records. The allowance level depends only on the input type and is fixed before any trajectory is generated, so it provides higher resolution where fine details matter without adapting image processing to outcomes. The processor resizes each complete image with its aspect ratio preserved, and admission uses the resulting token counts. Demonstration text and targets always stay intact, so a small context budget favors short records rather than truncated ones.
 
\noindent\textbf{Resource tiers.}
Table~\ref{tab:budgets} defines the three resource tiers. Training samples them jointly with count requests (Appendix~\ref{app:instance-construction}), and all main-text experiments use the Standard tier (Appendix~\ref{app:evaluation}).
 
\begin{table}[t]
\centering
\small
\setlength{\tabcolsep}{8pt}
\caption{Resource tiers. $H$ bounds the number of acquisition calls, $B_{\mathrm{acq}}$ bounds the cumulative input and generated tokens of all acquisition calls, and $B_{\mathrm{ctx}}$ bounds the retained demonstrations.}
\label{tab:budgets}
\begin{tabular}{@{}l r r r@{}}
\toprule
\textbf{Tier} & \textbf{Calls} $H$ & $B_{\mathrm{acq}}$ & $B_{\mathrm{ctx}}$ \\
\midrule
Tight    & 14 &  98,304 &  8,192 \\
Standard & 20 & 196,608 & 16,384 \\
Wide     & 28 & 393,216 & 32,768 \\
\bottomrule
\end{tabular}
\end{table}
 
\noindent\textbf{Commitment and fresh answering.}
Section~\ref{sec:method-context} defines admissibility and the commitment rule. We choose every $B_{\mathrm{ctx}}$ so that
\begin{equation}
    T(\text{ICL instructions})+T(q)+B_{\mathrm{ctx}}+G_{\mathrm{ans}}\le W,
\end{equation}
which guarantees that any admissible context fits a fresh answering call. An external answerer additionally requires the configured context budget to respect its own window. The environment never fills missing slots or replaces demonstrations on the agent's behalf, so every committed context reflects the agent's own choices. For $n=0$, the empty context is admissible from the start and a single \texttt{solve} commits it. In adaptive mode, the agent may also commit an empty context. The environment charges answering tokens separately from $B_{\mathrm{acq}}$, which keeps the acquisition cost tied to context construction.
 
\section{Training Data Construction}\label{app:training-data}
This appendix details how we build the 20,000 ICL instances of Section~\ref{sec:method-training}. We prepare and partition the sources (Appendices~\ref{app:source-preparation} and~\ref{app:source-grouping}), sample candidate configurations (Appendix~\ref{app:instance-construction}), and certify their feasibility with collection rollouts (Appendix~\ref{app:collection}).
 
\subsection{Source Preparation}\label{app:source-preparation}
We draw records from released training splits and from Meta-Dataset's meta-training classes, with a custom partition for BabyVision. Each record keeps its ordered images, original question and necessary input text, native answer options, and target output. We keep hidden OCR transcriptions, scene graphs, solving programs, generation code, and auxiliary explanations offline, so that demonstrations cannot supply supervision that is unavailable at test time. Table~\ref{tab:data-mixture} lists the allocation.
 
\noindent\textbf{Knowledge and commonsense VQA.}
We sample two- and three-hop ReasonVQA questions in a 60:40 target mixture and retain uniquely correct options without entity or knowledge-path annotations. We use multiple-choice A-OKVQA without rationales. For VCR Q$\rightarrow$A, we draw object references as consistently numbered boxes on the full image and omit rationale labels. All three sources are scored by multiple-choice accuracy.
 
\noindent\textbf{Text-rich image understanding.}
We retain single-page DocVQA images and questions without ground-truth OCR and score them with the released ANLS metric. For ChartQA, we emphasize human-written reasoning questions to cover comparison and aggregation, scored by relaxed accuracy. We render TabMWP tables as images so that the agent must read the visual input, and we apply the native answer normalization.
 
\noindent\textbf{Mathematical and diagram reasoning.}
We use corrected GeoQA diagrams, statements, and options in their original language, and we normalize units consistently for native option or numerical answers. We select image-bearing ScienceQA questions whose images participate in answering and exclude lecture and solution fields. AI2D keeps visible diagram labels but omits parsed annotations. ScienceQA and AI2D are scored by multiple-choice accuracy. From CoSyn, we select diagram and chart questions whose numerical, label, set, or sequence answers can be checked deterministically. To limit concentration on one visual format, we target at least six figure types in each CoSyn group and cap flowcharts at 20\% of the diagram group.
 
\noindent\textbf{Visual reasoning and native multi-image queries.}
We retain BabyVision tasks whose scoring rules reliably preserve coordinates, set structure, and order. NLVR2 keeps the left/right order of its photographs and is scored as binary statement verification. From Molmo2-SynMultiImageQA, we select two- to four-image questions that require information across images and admit a fixed answer parser, approximately balancing charts, tables, and diagrams. For CoSyn and Molmo2-SynMultiImageQA, we use the source data offline to verify that the rendered images contain the evidence required by the answer, and the agent never receives generation code or explanations. Omniglot, DTD, and CUB supply the visual concept-binding instances of Appendix~\ref{app:instance-construction}.
 
\noindent\textbf{Answer options.}
Before indexing, we shuffle the options of every multiple-choice record once with a record-specific seed and remap its answer, which reduces reliance on answer position. Thus, a record keeps the same permutation in every instance and rollout that uses it. Options whose meaning depends on their positions stay in place unless all references to them can be remapped consistently. We then precompute the embeddings from the final inputs, excluding targets and preparation metadata, so that retrieval scores correspond exactly to the records that the tools return.
 
\begin{table}[t]
\centering
\small
\setlength{\tabcolsep}{5pt}
\renewcommand{\arraystretch}{1.1}
\caption{Training mixture and demonstration-pool targets. RL counts are instances, and SFT counts are accepted trajectory--answer pairs per backbone. Bank counts are distinct records and are allocated independently of the query mixture.}
\label{tab:data-mixture}
\begin{tabular}{@{}l rrr r@{}}
\toprule
& \multicolumn{3}{c}{\textbf{Query mixture}} & \multicolumn{1}{c}{\textbf{Pool}} \\
\cmidrule(lr){2-4}\cmidrule(l){5-5}
\textbf{Source} & RL (\%) & RL & SFT & Bank \\
\midrule
ReasonVQA                & 10\phantom{.5} & 2,000 & 400 & 8,000 \\
A-OKVQA                  &  8\phantom{.5} & 1,600 & 320 & 5,000 \\
VCR (Q$\to$A)            &  8\phantom{.5} & 1,600 & 320 & 7,500 \\
DocVQA                   & 10\phantom{.5} & 2,000 & 400 & 5,500 \\
ChartQA                  &  8\phantom{.5} & 1,600 & 320 & 5,000 \\
TabMWP                   &  7\phantom{.5} & 1,400 & 280 & 5,000 \\
GeoQA                    &  8\phantom{.5} & 1,600 & 320 & 1,500 \\
ScienceQA (image subset) &  4\phantom{.5} &   800 & 160 & 1,000 \\
AI2D                     &  4\phantom{.5} &   800 & 160 &   800 \\
\addlinespace[2pt]
CoSyn                    &                &       &     &       \\
\quad diagrams           &  9\phantom{.5} & 1,800 & 360 & 4,000 \\
\quad charts             &  3\phantom{.5} &   600 & 120 & 1,500 \\
\addlinespace[2pt]
Meta-Dataset             &                &       &     &       \\
\quad Omniglot           &  6\phantom{.5} & 1,200 & 240 &   800 \\
\quad DTD                & 4.5            &   900 & 180 &   600 \\
\quad CUB                & 4.5            &   900 & 180 &   600 \\
\addlinespace[2pt]
BabyVision               &  1\phantom{.5} &   200 &  40 &   208 \\
NLVR2                    &  3\phantom{.5} &   600 & 120 & 1,200 \\
Molmo2-SynMultiImageQA   &  2\phantom{.5} &   400 &  80 & 1,800 \\
\midrule
\textbf{Total}           & \textbf{100}\phantom{.5} & \textbf{20,000} & \textbf{4,000} & \textbf{50,008} \\
\bottomrule
\end{tabular}
\end{table}
 
\subsection{Deduplication and Pool Partitioning}\label{app:source-grouping}
Before sampling, we remove exact duplicates, namely records whose images, input text, and target all coincide, including copies shared across sources. Records that share only an image or only a question remain distinct, because a different question about the same image, or the same question about a different image, poses a different case. We then assign each remaining record to exactly one of the demonstration pool, the training-query pool, or the development pool while preserving task-subtype coverage where possible. Thus, no query reappears as a bank record, although a query and a bank record may share an image or a question. The 1,000 development queries are disjoint from both training queries and bank records, which lets us diagnose the policy without touching evaluation benchmarks. BabyVision releases 388 questions. We partition them by subtype into 208 bank records, 140 training queries, and 40 development queries. We use 100 of the training queries to form 200 RL instances and target 40 accepted SFT pairs.
 
\subsection{ICL Instance Construction}\label{app:instance-construction}
Each instance fixes its query, bank, label or option mapping, image processing, count request, and resource tier before any rollout, so that all comparisons within an RL group reflect acquisition decisions under an identical task. The sampler pairs each sampled query with two configurations that differ in bank composition or resource request.
 
\noindent\textbf{Demonstration-bank sampling.}
The sampler builds each static bank from a 256-record core and a background. It first draws the bank size from $\{512, 1024, 4096\}$ with probabilities 20\%, 50\%, and 30\%. It then fills the core under a profile drawn from Table~\ref{tab:bank-mixture} and completes the bank to its sampled size by drawing uniformly without replacement from all unselected records of the mixed pool, including the query's own source. Fixing the core size lets larger banks enlarge the background without changing the route-controlled portion. Varying the profile exposes the agent to different mixtures of task-matched guidance, similar inputs, and cross-source examples. The three routes of Section~\ref{sec:method-training} fill the core in order as follows.
\begin{enumerate}
    \item \textbf{Source-matched sampling.} The sampler draws uniformly from the query's released task subset and falls back to the whole source pool when no subdivision exists. The manifest records every subset definition, and membership in a training or test split never counts as a task subtype. For visual concept binding, the sampler draws from the instance's selected classes with approximately balanced counts, so that each label convention has examples.
    \item \textbf{Input-similarity retrieval.} The sampler supplies the query images as reference images and the query's input text, when present, to both text channels. It reuses the encoders and fusion of Appendix~\ref{app:retrieval} but ranks the full eligible pool without the cutoff $N_{\mathrm{list}}$, because a core requires up to 160 similar records under its profile quota, and more when a source-matched shortfall is transferred to this route. It takes the highest-ranked unselected records. The sampler uses neither the query target nor task metadata, so similar records may still differ in instructional value. The agent never sees these offline rankings.
    \item \textbf{Random cross-source sampling.} The sampler draws uniformly from eligible records outside the query's source, without conditioning on similarity or metadata, to provide background beyond the first two routes.
\end{enumerate}
The sampler removes each selected record from every remaining route pool to prevent duplicates. When the source-matched quota cannot be filled, it transfers the shortfall to similarity retrieval and then, if needed, to cross-source sampling. It logs each record's route together with the target and realized quotas, which makes fallback behavior reproducible, and keeps these logs hidden from the agent because they describe sampling provenance instead of demonstration utility. We impose no separate redundancy quota and assign no relevance or distractor labels, since distinct records can overlap in guidance within any route. Training instead teaches the agent to judge what each demonstration adds through downstream answer quality, and the agent discovers candidates only through its own searches. 
 
\begin{table}[t]
\centering
\small
\setlength{\tabcolsep}{5pt}
\caption{Sampling profiles for the 256-record core. Route percentages apply within the core, and the last column gives the probability of selecting each profile.}
\label{tab:bank-mixture}
\begin{tabular}{@{}l ccc c@{}}
\toprule
& \multicolumn{3}{c}{\textbf{Core routes (\%)}} & \\
\cmidrule(lr){2-4}
\textbf{Profile} & Source & Similarity & Cross & \textbf{Prob.\ (\%)} \\
\midrule
Source-heavy     & 50\phantom{.5} & 25\phantom{.5} & 25 & 20 \\
Mixed            & 25\phantom{.5} & 50\phantom{.5} & 25 & 50 \\
Similarity-heavy & 12.5           & 62.5           & 25 & 30 \\
\bottomrule
\end{tabular}
\end{table}
 
\noindent\textbf{Visual concept binding.}
For the 15\% of instances drawn from Meta-Dataset, the sampler selects two to six classes and assigns them random neutral output labels. The query lists the allowed labels without revealing their class mapping, so the agent can infer the convention only from demonstrations. One mapping holds for all selected classes within an instance and changes across instances, while the indexed images and generic classification inputs stay fixed, so the retrieval embeddings are unaffected by the mapping. Class identifiers and mapping tables are withheld from the agent, and predicted labels are scored by exact match. 
 
\noindent\textbf{Count requests and resource tiers.}
The sampler draws the count request and the resource tier jointly from Table~\ref{tab:shot-budget} once per instance. Half of the instances use adaptive acquisition, and joint sampling assigns demanding fixed counts mainly to the Standard and Wide tiers, where a feasible path is more likely to fit. We do not sample $n=0$ as a fixed-count category, although an adaptive agent may commit an empty context. 
Before any rollout, the sampler rejects a fixed-count configuration when the $n$ shortest complete demonstrations in its bank exceed $B_{\mathrm{ctx}}$, or when $H<\lceil n/N_{\mathrm{cand}}\rceil+n+2$, which lower-bounds the calls needed for enough searches, one \texttt{read} per demonstration, one \texttt{curate}, and one \texttt{solve}. These checks cannot foresee the growing observation history, so execution validates the remaining configurations.
 
\begin{table}[t]
\centering
\small
\setlength{\tabcolsep}{6pt}
\def\hh{\phantom{.5}}%
\caption{Joint distribution of count requests and resource tiers, as percentages of all training instances. A configuration is admitted only after a feasible acquisition path under its limits has been validated.}
\label{tab:shot-budget}
\begin{tabular}{@{}l r rrrr r@{}}
\toprule
& & \multicolumn{4}{c}{\textbf{Fixed count} $n$} & \\
\cmidrule(lr){3-6}
\textbf{Tier} & \textbf{Adaptive} & 1 & 2 & 4 & 8 & \textbf{Total} \\
\midrule
Tight    & 10 & 2.5  & 4 & 2.5  & 1  & 20 \\
Standard & 25 & 1.5  & 7 & 11 & 5.5  & 50 \\
Wide     & 15 & 1 & 4 & 6.5  & 3.5  & 30 \\
\midrule
\textbf{Total} & \textbf{50} & \textbf{5}\hh & \textbf{15} & \textbf{20}\hh & \textbf{10}\hh & \textbf{100} \\
\bottomrule
\end{tabular}
\end{table}
 
\subsection{Collection Rollouts and Instance Acceptance}\label{app:collection}
Each backbone's frozen instruction-tuned checkpoint executes four acquisition attempts per candidate configuration through the actual tools, sampling with temperature 0.8 and top-$p$ 0.95. For every committed admissible context, the same checkpoint generates one fresh answer. These collection rollouts both certify that each configuration is feasible and supply the SFT candidates of Appendix~\ref{app:sft}, which avoids a separate exploration stage.
 
\noindent\textbf{Path validation.}
A validator replays every call of a recorded path against Eq.~\ref{eq:appendix-path-check} with complete serialized observations. It checks actions, indices, state transitions, the final count, the context size, and the answering reservation of Appendix~\ref{sec:acquisition-protocol}. Visual and text token counts differ across backbones, so the validator accepts a configuration only when one recorded path to an admissible context replays validly under every backbone's processor and chat template. The accepted configurations thus form a manifest shared by all backbones. For adaptive concept-binding instances, the validator additionally verifies offline that the support information is reachable through the tools. This dataset-level check neither requires the runtime agent to retain particular examples nor changes the admissibility rule. Path records stay hidden from the policy.
 
\noindent\textbf{Acceptance and replacement.}
Acceptance depends on the path rather than on the answer, so an incorrect initial completion alone never excludes a feasible configuration from RL. The sampler replaces each configuration without a feasible path by a new draw from the same source and request/tier stratum. If a stratum runs short, it draws from the same capability group where possible while keeping 15\% concept-binding instances and at most 1\% BabyVision. We record realized counts instead of duplicating records or trajectories to fill quotas. The final manifest contains 20,000 instances over approximately 10,000 distinct queries. Screening removes only configurations that admit no feasible path, and RL later retains every policy failure on accepted instances (Appendix~\ref{sec:rl-details}).
 
\section{Rejection-Sampling SFT}\label{app:sft}
\noindent\textbf{Pair selection.}
Each backbone selects SFT data from its own collection rollouts on the accepted manifest, which keeps supervision within the behavior that the backbone can already reach. A trajectory--answer pair qualifies when all calls are valid, resource use is compliant, the committed context is admissible, and the answer receives the full score under the source metric. Each backbone keeps 4,000 qualifying pairs with at most two per query, following the SFT targets of Table~\ref{tab:data-mixture} and covering sources, count requests, resource tiers, and action patterns.
 
\noindent\textbf{Objective.}
For a sequence $z$ generated over one or more model calls, let $\mathrm{gen}(z)$ index its generated tokens. For $k\in\mathrm{gen}(z)$, let $\xi_k$ denote the complete input of the call that produced $z_k$, followed by the tokens generated earlier within that call. An acquisition call's input is the serialized observation, so reasoning replayed from earlier calls appears in $\xi_k$ only as conditioning and is never supervised twice. Over the SFT set $\mathcal S$ of pairs $(\tau,\hat y)$, the objective of Section~\ref{sec:method-sft} reads
\begin{equation}
    \mathcal L_{\mathrm{SFT}}(\theta)=
    -\frac{1}{|\mathcal S|}\sum_{(\tau,\hat y)\in\mathcal S}\,
    \sum_{z\in\{\tau,\hat y\}}\,\sum_{k\in\mathrm{gen}(z)}
    \log\pi_\theta(z_k\mid\xi_k).
    \label{eq:sft-loss}
\end{equation}
We implement SFT with LLaMA-Factory~\citep{zheng2024llamafactory} 
and extend it with complete-pair batching, so every minibatch contains whole pairs and estimates Eq.~\ref{eq:sft-loss} with the same reduction.
 
\section{Joint Reinforcement Learning}\label{sec:rl-details}
\noindent\textbf{Rollout groups.}
Each backbone starts RL from its own SFT checkpoint and samples fresh on-policy rollouts over the 20,000 accepted instances instead of reusing collection trajectories. At the start of every rollout batch, the trainer freezes the current parameters as $\theta_{\mathrm{old}}$. For each instance in the batch, $\pi_{\theta_{\mathrm{old}}}$ samples $K$ trajectories $\tau_1,\dots,\tau_K$ under the instance's count request and budgets. For each valid trajectory, the same policy independently samples $M$ answers $\hat y_{i1},\dots,\hat y_{iM}$ in fresh calls whose input consists of the ICL instructions, $q$, and $S_i$. A failed trajectory ($\delta_i=0$) receives no answers. We call the $K$ trajectories together with all their answers a group. One frozen policy generates both acquisition and answering tokens, so every likelihood ratio in the batch refers to the policy that produced the token. We require $K\ge2$ and $M\ge2$ because both advantages below use leave-one-out baselines, and we keep $K$ and $M$ fixed across groups.
 
\noindent\textbf{Return bounds.}
Consider a valid trajectory under the return of Eq.~\ref{eq:acquisition-return}. Admissibility gives $T(S_i)\le B_{\mathrm{ctx}}$, and admission gives $\sum_t(I_{it}+O_{it})\le B_{\mathrm{acq}}$ (Appendix~\ref{sec:acquisition-protocol}). Both normalized costs lie in $[0,1]$, and since $\bar r_i\in[0,1]$ and $\lambda_{\mathrm{ctx}},\lambda_{\mathrm{acq}}\ge0$,
\begin{equation}
    -\lambda_{\mathrm{ctx}}-\lambda_{\mathrm{acq}}\le R_i\le 1
    \qquad\text{whenever }\delta_i=1.
\end{equation}
Any $\kappa>\lambda_{\mathrm{ctx}}+\lambda_{\mathrm{acq}}$ thus makes every valid return strictly larger than the failure return $-\kappa$. Failed trajectories remain in their groups, so the policy learns from its own acquisition failures.
 
\noindent\textbf{Cost schedule.}
The trainer lets the policy first learn to commit admissible contexts and only then pushes it toward efficiency. It scales both cost weights by a common ramp,
\begin{equation}
    \begin{gathered}
    \lambda_{\mathrm{ctx}}(\zeta)=\lambda_{\mathrm{ctx}}^{\mathrm{fin}}\,g(\zeta),\qquad
    \lambda_{\mathrm{acq}}(\zeta)=\lambda_{\mathrm{acq}}^{\mathrm{fin}}\,g(\zeta),\\
    g(\zeta)=\min\Big\{1,\max\Big\{0,\frac{\zeta-\zeta_0}{\zeta_1-\zeta_0}\Big\}\Big\},
    \end{gathered}
    \label{eq:cost-schedule}
\end{equation}
where $\zeta\in[0,1]$ is the fraction of the 20,000 instances sampled before the current rollout batch, $\zeta_0=0.1$, and $\zeta_1=0.3$. The weights stay fixed within a rollout batch, so all returns of a batch share one reward definition. Because $g(\zeta)\le1$, the condition $\kappa>\lambda_{\mathrm{ctx}}^{\mathrm{fin}}+\lambda_{\mathrm{acq}}^{\mathrm{fin}}$ ensures the bound above throughout training, and our values in Table~\ref{tab:rl-parameters} give $1>0.07$.
 
\noindent\textbf{Advantages.}
The trainer compares each acquisition with the other $K-1$ acquisitions of the same instance, which evaluates a path against alternatives under an identical request and identical resources~\citep{rloo}:
\begin{equation}
    A_i^{\mathrm{acq}}=R_i-\frac{1}{K-1}\sum_{i'\ne i}R_{i'}.
    \label{eq:acquisition-advantage}
\end{equation}
The baseline includes failed trajectories, and the advantages of a group sum to zero. For answering, Section~\ref{sec:method-rl} defines $A_{ij}^{\mathrm{ans}}$ with the leave-one-out mean
\begin{equation}
    \bar r_{i,-j}=\frac{1}{M-1}\sum_{j'\ne j}r_{ij'}.
\end{equation}
The trainer computes both advantages once per rollout batch from samples of $\pi_{\theta_{\mathrm{old}}}$ and treats them as constants with respect to $\theta$. It applies no further group centering or standard-deviation normalization, because rescaling would alter the $\pm\mu/2$ signal of uniform answer groups and inflate advantages in nearly uniform groups.
 
\noindent\textbf{Clipped surrogate.}
With $\mathrm{gen}(z)$ and $\xi_k$ as in Appendix~\ref{app:sft}, the likelihood ratio of token $z_k$ is
\begin{equation}
    \rho_k(\theta)=\frac{\pi_\theta(z_k\mid\xi_k)}{\pi_{\theta_{\mathrm{old}}}(z_k\mid\xi_k)},
\end{equation}
whose denominator is stored at rollout time. For a sequence $z$ and a scalar advantage $A$, the surrogate in Eq.~\ref{eq:joint-policy-loss} is
\begin{equation}
    \Phi_\theta(z,A)=\sum_{k\in\mathrm{gen}(z)}
    \min\big\{\rho_k(\theta)A,\;
    \operatorname{clip}\big(\rho_k(\theta),1-\epsilon,1+\epsilon\big)A\big\}.
    \label{eq:clipped-sequence}
\end{equation}
For a trajectory $\tau_i$, $\mathrm{gen}(\tau_i)$ collects the generated tokens of all its acquisition calls, including calls that the environment rejected, because the policy produced them and their cost entered $R_i$. All tokens of $\tau_i$ share $A_i^{\mathrm{acq}}$, and all tokens of $\hat y_{ij}$ share $A_{ij}^{\mathrm{ans}}$. Instructions, images, tool outputs, and replayed history serve only as conditioning, so replay creates no additional loss targets. When evaluating $\pi_\theta$, the trainer reconstructs each call's serialized input and action constraints exactly as during rollout, so the numerator and denominator of $\rho_k$ condition on the same $\xi_k$.
 
\noindent\textbf{Loss normalization.}
Eq.~\ref{eq:joint-policy-loss} divides by the fixed constant $L_{\mathrm{ref}}$ instead of sequence length, which avoids inversely weighting the tokens of longer trajectories~\citep{liu2025understanding}. The answer term keeps the denominator $KM$ even when fewer than $K$ contexts are valid. Therefore, every sampled answer carries the same coefficient, and answers from groups with many failed acquisitions receive no extra weight.
 
\noindent\textbf{Optimization schedule.}
After collecting and scoring a rollout batch of 32 instance groups, the trainer shuffles the groups into four minibatches of eight complete groups and performs one update per minibatch in sequence. Each update estimates the expectation in Eq.~\ref{eq:joint-policy-loss} by averaging the bracketed term over the eight groups of its minibatch. The snapshot $\theta_{\mathrm{old}}$, the stored log-probabilities, the returns, and the advantages stay fixed across the four updates and are refreshed only before the next rollout batch, so every ratio compares against the policy that generated the batch. One pass over the manifest thus yields 625 rollout batches and 2,500 updates, with 80,000 acquisitions and at most 160,000 fresh answers per backbone beyond the collection rollouts. Microbatching and gradient accumulation leave this reduction unchanged because minibatches contain complete groups. We implement this loop in VERL~\citep{verl}.
The objective contains no KL, entropy, format, retrieval-hit, action, or MoE-routing auxiliary term, which keeps every update tied to the answer and resource objective.

\section{Implementation Details}\label{app:implementation}

We instantiate \method{} with Qwen3-VL-8B-Instruct, Qwen3-VL-30B-A3B-Instruct, and Qwen3.5-9B.
All three backbones use their native processors and chat templates and share parameters between acquisition and answering. We update all language-backbone parameters, including the language MoE parameters of Qwen3-VL-30B-A3B, and freeze the visual encoders, mergers, and projections to retain the pretrained visual representation. Both training stages use AdamW and run on 8 NVIDIA H200 GPUs. Tables~\ref{tab:optimization} and~\ref{tab:rl-parameters} list all training settings. Each run stores its tool schemas, index mappings, processor settings, and prompt templates (Appendix~\ref{app:prompt}).
 
\begin{table}[t]
\centering
\small
\setlength{\tabcolsep}{5pt}
\renewcommand{\arraystretch}{1.15}
\caption{Training settings per backbone. A joint-RL group contains all $K$ acquisition trajectories of an instance and the answers sampled from its valid contexts.}
\label{tab:optimization}
\begin{tabularx}{\linewidth}{@{}>{\raggedright\arraybackslash}p{0.34\linewidth} >{\raggedright\arraybackslash}X >{\raggedright\arraybackslash}X@{}}
\toprule
\textbf{Setting} & \textbf{RS-SFT} & \textbf{Joint RL} \\
\midrule
\multicolumn{3}{@{}l}{\textit{Training volume and batching}} \\
\addlinespace[2pt]
Training volume & 4,000 pairs, 2 epochs & 20,000 instances, 1 pass \\
Global batch & 16 pairs & 32 groups per rollout batch \\
Optimizer minibatch & 16 pairs & 8 complete groups \\
Optimizer updates & 500 & 2,500 \\
\addlinespace[5pt]
\multicolumn{3}{@{}l}{\textit{Optimizer and numerics}} \\
\addlinespace[2pt]
Optimizer & AdamW & AdamW \\ 
Learning rate & $10^{-5}$ & $5\times10^{-7}$ \\
Learning-rate schedule & 15-update warmup, cosine decay & 75-update warmup, then constant \\
Weight decay & 0.01 & 0 \\
Precision & BF16 & BF16 \\
Gradient-norm clipping & 1.0 & 1.0 \\
\addlinespace[5pt]
\multicolumn{3}{@{}l}{\textit{Rollout decoding}} \\
\addlinespace[2pt]
Temperature & 0.8 (collection) & 1.0 \\
Top-$p$ & 0.95 (collection) & 1.0 \\
Top-$k$ and min-$p$ truncation & Disabled & Disabled \\
\bottomrule
\end{tabularx}
\end{table}
 
\begin{table}[t]
\centering
\small
\setlength{\tabcolsep}{5pt}
\caption{Joint RL hyperparameters. The cost schedule depends on the fraction of instances sampled and is independent of the learning-rate schedule.}
\label{tab:rl-parameters}
\begin{tabular}{@{}l c l@{}}
\toprule
\textbf{Parameter} & \textbf{Symbol} & \textbf{Value} \\
\midrule
Acquisition trajectories per instance & $K$ & 4 \\
Answers per valid context & $M$ & 2 \\
\addlinespace[3pt]
Clipping width & $\epsilon$ & 0.2 \\
Answer baseline mixture & $\mu$ & 0.1 \\
Answer reference baseline & $b_{\mathrm{ref}}$ & 0.5 \\
Fixed loss normalizer & $L_{\mathrm{ref}}$ & 8,192 \\
\addlinespace[3pt]
Failure penalty & $\kappa$ & 1 \\
Final context cost weight & $\lambda_{\mathrm{ctx}}^{\mathrm{fin}}$ & 0.05 \\
Final acquisition cost weight & $\lambda_{\mathrm{acq}}^{\mathrm{fin}}$ & 0.02 \\
Cost ramp endpoints & $(\zeta_0,\zeta_1)$ & $(0.1,\,0.3)$ \\
\addlinespace[3pt]
KL coefficient & -- & 0 \\
Entropy bonus & -- & 0 \\
\bottomrule
\end{tabular}
\end{table}
 
\section{Evaluation Protocol}\label{app:evaluation}
\subsection{Benchmark Pools and Partitions}
For VL-ICL Bench, we evaluate its image-to-text tasks. Before constructing support and query pools, we check every candidate record against the frozen training-bank, training-query, and development manifests, so that no evaluation case repeats a training or development case. We remove exact duplicates under the criterion of Appendix~\ref{app:source-grouping} from the evaluation side, including evaluation-bank records that a benchmark releases as a training split. 
 
\begin{table}[t]
\centering
\small
\setlength{\tabcolsep}{5pt}
\renewcommand{\arraystretch}{1.15}
\caption{Evaluation pools and stratification. Benchmarks with released splits keep their native bank\,/\,query boundaries. The others receive a new partition that targets a 6:4 bank\,/\,query ratio within each stratum.}
\label{tab:evaluation-splits}
\begin{tabularx}{\linewidth}{@{}l >{\raggedright\arraybackslash}X >{\raggedright\arraybackslash}X@{}}
\toprule
\textbf{Benchmark} & \textbf{Available pool (bank\,/\,query)} & \textbf{Stratification} \\
\midrule
\multicolumn{3}{@{}l}{\textit{Released splits preserved}} \\
\addlinespace[2pt]
TextVQA & train\,/\,validation & native boundary \\
\addlinespace[3pt]
GVL & train\,/\,test & native boundary, per graph task \\
\addlinespace[3pt]
VL-ICL Bench & released image-to-text support\,/\,query construction & per subtask \\
\midrule
\multicolumn{3}{@{}l}{\textit{New 6:4 partition}} \\
\addlinespace[2pt]
MME-Reasoning & public labeled release & reasoning category and subtype, difficulty where available \\
\addlinespace[3pt]
CharXiv & public labeled validation set & question type (descriptive\,/\,reasoning), chart category \\
\addlinespace[3pt]
MathVision & full public release (3{,}040 problems) & subject, difficulty \\
\addlinespace[3pt]
MDK12 & image-bearing records from the official mini easy\,/\,medium\,/\,hard archives & discipline, subtype, difficulty \\
\bottomrule
\end{tabularx}
\end{table}
 
\noindent\textbf{Stratified partitioning.}
For benchmarks without released splits, we target 60\% supports and 40\% queries, which keeps a majority of records for retrieval while leaving enough queries for evaluation. We assign individual records with seed 42, stratifying by released subcategory and balancing difficulty within each subtype where feasible, so that every stratum keeps both supports and queries. Sparse secondary strata are balanced at their parent category, and reported subtype labels remain unchanged. Rounding within small strata can prevent an exact 6:4 ratio, so we record realized counts. For every benchmark, including those with released splits, we also remove exact duplicates across the bank\,/\,query boundary. The same procedure gives the 6:4 partitions of MMStar, PerceptionBench, MindCube, and MMMU-Pro used in Appendix~\ref{app:eval-inference}.
 
After partitioning, we merge the eligible supports of all subcategories into one bank per benchmark, subject to native task compatibility. The agent receives no query-subcategory filter, so stratification controls evaluation coverage without revealing which examples to retrieve. These banks carry no training-route quotas or cross-source additions. CharXiv uses its labeled validation set because public test labels are unavailable. MathVision's testmini subset is contained in the full release, so we do not add it as a separate support source. The manifest records MDK12 archive revisions and retained records, and we report all custom ICL partitions separately from full-set leaderboard settings.
 
\noindent\textbf{Original task setting.}
VL-ICL keeps each subtask's admissible support construction and instance-specific label or operator convention. We count individual input--output pairs as demonstrations, so when a native VL-ICL routine supplies two class-specific supports per shot, each native shot corresponds to two \method{} demonstrations. Comparisons match the number of examples actually supplied and use only counts that the task supports. 
 
\subsection{Baselines}\label{app:baselines}
All acquisition baselines draw from the same evaluation banks and share the query partitions, image processing, backbone architectures, and answering prompt of \method{}.
\begin{itemize}
    \item \textbf{Random sampling} draws demonstrations uniformly from the bank.
    \item \textbf{MMICES}
    selects demonstrations by multimodal similarity with the same SigLIP encoder as Appendix~\ref{app:retrieval}.
    \item \textbf{ContextNav}
    uses Gemini3.5-Flash as its policy and passes the constructed context to the answering model.
    \item \textbf{LSD}
    trains a retriever with answering-model feedback, using the same data sources as \method{} and its original data construction method.
    \item \textbf{ProMSA}
    is adapted to multimodal ICL by running it in the same environment as \method{} (Appendix~\ref{sec:tool-protocol}). We train it with SFT and RL on the same data as \method{} at a comparable training cost, and it answers with its own trained model, ensuring a fair comparison.
\end{itemize}
 
\noindent\textbf{Fine-tuned baselines for supplied contexts.}
We use 8-rank LoRA and DARA to fine-tune the same backbone checkpoints as \method{} to improve interleaved image--text understanding. Their few-shot training data come from the same sources as \method{} (Table~\ref{tab:data-mixture}). For each training query, we build 4-shot and 8-shot interleaved sequences by retrieving demonstrations from the demonstration pool with the fused image--text similarity of the input-similarity route, and supervise the query's ground-truth target. We further add COMM~\citep{comm} to strengthen multi-image understanding. Because supervised fine-tuning and RL use data differently, we do not equate the two training volumes. Instead, we train each baseline at six data sizes capped at the \method{} training volume and report the size that has the best performance. The selected configuration uses about 35\% of the \method{} training data.
 
\subsection{Scoring and Inference}\label{app:eval-inference}
We apply each benchmark's released answer extraction and scoring, namely multi-reference VQA accuracy for TextVQA, task-specific extraction and judging for MME-Reasoning, descriptive and reasoning scores for CharXiv, graph-task scoring for GVL, mathematical answer scoring for MathVision, discipline-specific scoring for MDK12, and native image-to-text metrics for VL-ICL. Benchmarks that require an external judge use its released configuration, only for evaluation. We report each benchmark’s default metrics.
 
\noindent\textbf{Acquisition and answering.}
All main-text experiments use the Standard tier of Table~\ref{tab:budgets}, and every main-experiment result is the mean over three random seeds. Fixed-count baselines use four demonstrations unless a figure varies $n$. At inference, the agent acquires greedily and constructs one context, after which the answering model generates one greedy answer in a fresh call. Query targets reach only the scorer. Acquisition failures stay in the denominator with zero task score, so that the evaluation does not condition on successful acquisition.
 
\noindent\textbf{Answering with supplied contexts.}
To isolate the ability to use context, we disable acquisition and give each model fixed inputs. For MME-Reasoning, MathVision, and MMStar, the base backbone, LoRA, DARA, and the jointly trained \method{} checkpoint receive identical 4- and 8-shot prompts constructed by MMICES on the same partitions. MuirBench and BLINK present each case as an interleaved image--text sequence, which we evaluate as released. \method{} uses no separate training branch for supplied contexts. 
 
\noindent\textbf{Acquisition for external answerers.}
The \method{}-trained Qwen3.5-9B constructs 4-shot contexts for the external answering models on PerceptionBench, MindCube, and MMMU-Pro, acquiring demonstrations from each benchmark's support bank. For each answerer, we fix the checkpoint or API version and the answering configuration across \method{}, MMICES, and ContextNav, so that score differences reflect context construction.

\section{Additional ablation studies and analyses}\label{app:ablations}
\begin{table}[h]
\centering
\small
\setlength{\tabcolsep}{5pt}
\caption{Ablation on training stages with Qwen3.5-9B under four-shot requests. ``w/o acquisition'' and ``w/o answering'' run joint RL after SFT without the corresponding part.}
\label{tab:ablation-stages-4shot}
\begin{tabular}{@{}l ccc c@{}}
\toprule
\textbf{Method} & \textbf{MME-R} & \textbf{CharXiv} & \textbf{MathVision} & \textbf{Avg.} \\
\midrule
Base            & 0.343 & 0.675 & 0.728 & 0.582 \\
SFT only        & \textbf{0.431} & 0.743 & 0.854 & 0.676 \\
Joint RL only   & 0.423 & 0.748 & 0.850 & 0.674 \\
\midrule
w/o acquisition & 0.402 & 0.740 & 0.838 & 0.660 \\
w/o answering   & 0.415 & 0.745 & 0.847 & 0.669 \\
\method{}       & 0.427 & \textbf{0.751} & \textbf{0.862} & \textbf{0.680} \\
\bottomrule
\end{tabular}
\end{table}

\textbf{Training stages under four-shot requests.} Table~\ref{tab:ablation-stages-4shot} complements the training-stage ablation with four-shot requests, which remove the agent's control over the number of demonstrations. Either training stage alone again brings large gains over the base checkpoint, raising the average from 0.582 to 0.676 with SFT and to 0.674 with joint RL, and the full model remains best on average at 0.680. Two differences from the adaptive setting stand out. First, joint RL after SFT adds only 0.4 percentage points over SFT alone, compared with 5.2 points under the adaptive setting, and SFT alone is even slightly better on MME-Reasoning. Thus, most of the gain from joint RL comes from learning how many demonstrations to commit, a decision that fixed-count requests take away. Second, optimizing only one part after SFT falls below SFT alone, to 0.660 without acquisition and 0.669 without answering. Because acquisition and answering share parameters, updating one part alone disturbs the other, and only optimizing both together improves on SFT. Removing acquisition lowers the average by 2.0 points, as in the adaptive setting, whereas removing answering lowers it by only 1.1 points, since a fixed set of four demonstrations leaves less room to improve how the agent answers from them.

\begin{wraptable}{r}{0.6\textwidth}
\vspace{-12pt}
\centering
\footnotesize
\setlength{\tabcolsep}{3.5pt}
\caption{Ablation on \texttt{read} and \texttt{curate} with Qwen3.5-9B under the adaptive setting.}
\label{tab:ablation-actions}
\begin{tabular}{@{}l ccc c@{}}
\toprule
\textbf{Variant} & \textbf{MME-R} & \textbf{CharXiv} & \textbf{MathVision} & \textbf{Avg.} \\
\midrule
w/o \texttt{read}   & 0.445 & 0.758 & 0.852 & 0.685 \\
w/o \texttt{curate} & 0.448 & 0.771 & 0.859 & 0.693 \\
Add-only curate     & 0.457 & 0.782 & 0.862 & 0.700 \\
\method{}           & \textbf{0.468} & \textbf{0.788} & \textbf{0.870} & \textbf{0.709} \\
\bottomrule
\end{tabular}
\vspace{-10pt}
\end{wraptable}
\textbf{\texttt{read} and \texttt{curate} actions.} \texttt{read} and \texttt{curate} separate what the agent sees from what it keeps. To test both, we restrict the tool set of the trained agent at inference in three ways. Without \texttt{read}, \texttt{search} returns complete demonstrations instead of previews. Without \texttt{curate}, every read demonstration joins the context in read order until the count or context budget is reached, and nothing can be removed. Add-only curate keeps additions but forbids removals, which isolates the value of revising the context. 

Table~\ref{tab:ablation-actions} shows that every restriction lowers the score on all three benchmarks, and that the three variants form a clear ordering. Removing \texttt{read} hurts most, lowering the average from 0.709 to 0.685. Without previews, every retrieved candidate enters later observations in full, so the acquisition budget runs out after fewer searches and the agent compares fewer alternatives before committing. The loss is largest on CharXiv, at 3.0 percentage points, whose dense chart images are the most expensive inputs to inspect in full. Removing \texttt{curate} lowers the average to 0.693, with the largest loss of 2.0 points on MME-R. The agent still screens candidates through previews, but it must keep every demonstration it opens, including those that prove unhelpful once their targets become visible. Add-only curate recovers part of this gap and reaches 0.700, yet it still trails the full model by 0.9 points on average and by 1.1 points on MME-R. Of the 1.6 points gained through curation, choosing what to add therefore accounts for 0.7 points, whereas the ability to remove demonstrations accounts for the remaining 0.9 points. Trajectory statistics confirm that the agent relies on both abilities in practice. Across all main experiments with the three backbones, it retains 74.2\% of the demonstrations it reads and removes 0.85 demonstrations per query on average. Together, these findings demonstrate the necessity of our action-space design.

\begin{table}[h]
\centering
\small
\setlength{\tabcolsep}{4pt}
\caption{Reward ablation with Qwen3.5-9B under the adaptive setting. \#Shots is the mean committed count and Acq.\ is the mean number of acquisition tokens, both averaged over the three benchmarks.}
\label{tab:ablation-reward}
\begin{tabular}{@{}l ccc c cc@{}}
\toprule
\textbf{Variant} & \textbf{MME-R} & \textbf{CharXiv} & \textbf{MathVision} & \textbf{Avg.} & \#Shots & Acq.\,(k) \\
\midrule
w/o resource costs  & 0.461 & 0.779 & 0.868 & 0.703 & 2.94 & 32.7 \\
w/o cost schedule   & 0.459 & 0.764 & 0.857 & 0.693 & 2.78 & 30.2 \\
w/o fixed reference & 0.442 & 0.779 & 0.857 & 0.693 & 2.10 & 28.9 \\
\method{}           & \textbf{0.468} & \textbf{0.788} & \textbf{0.870} & \textbf{0.709} & 2.62 & 26.3 \\
\bottomrule
\end{tabular}
\end{table}

\textbf{Reward design.} We ablate three design choices of the RL objective in Section~\ref{sec:method-rl}, retraining each variant from the same SFT checkpoint with all other settings fixed. We remove resource costs by setting $\lambda_{\mathrm{ctx}}=\lambda_{\mathrm{acq}}=0$, remove the cost schedule by applying the final cost weights from the first rollout batch, and remove the fixed reference by setting $\mu=0$. Since the costs target efficiency, Table~\ref{tab:ablation-reward} also reports the committed count and acquisition tokens. Removing resource costs lowers the average only slightly, from 0.709 to 0.703, but the agent commits 2.94 demonstrations instead of 2.62 and spends 24\% more acquisition tokens. The costs thus make acquisition more economical without sacrificing accuracy. Applying the costs from the start instead lowers the average to 0.693 and is also less efficient, with 15\% more acquisition tokens. Before the policy reliably commits admissible contexts, immediate cost pressure only adds penalties to returns that still often end in failure, so the schedule is needed to introduce the costs stably.
Removing the fixed reference also lowers the average to 0.693, with the largest loss of 2.6 percentage points on MME-R. With binary rewards and two answers per context, many answer groups are uniformly correct or incorrect, and a pure leave-one-out baseline gives them zero advantage. This is most common on MME-R, where accuracy is lowest, and the fixed reference keeps a signal of $\pm\mu/2$ in such groups. Overall, the resource costs mainly improve efficiency, whereas the cost schedule and the fixed reference each contribute about 1.5 points in accuracy.

\begin{table}[t]
\centering
\small
\setlength{\tabcolsep}{5pt}
\caption{Ablation on training-bank construction with Qwen3.5-9B under the adaptive setting. The top block evaluates each benchmark on its own support bank, and the bottom block on a mixed bank that merges the support banks of the three benchmarks.}
\label{tab:ablation-bank}
\begin{tabular}{@{}l ccc c@{}}
\toprule
\textbf{Training bank} & \textbf{MME-R} & \textbf{CharXiv} & \textbf{MathVision} & \textbf{Avg.} \\
\midrule
\multicolumn{5}{@{}l}{\textit{Own bank}} \\
\addlinespace[2pt]
Source-only & 0.459 & 0.767 & 0.862 & 0.696 \\
Random      & 0.448 & 0.759 & 0.858 & 0.688 \\
\method{}   & \textbf{0.468} & \textbf{0.788} & \textbf{0.870} & \textbf{0.709} \\
\midrule
\multicolumn{5}{@{}l}{\textit{Mixed bank}} \\
\addlinespace[2pt]
Source-only & 0.422 & 0.743 & 0.844 & 0.670 \\
Random      & 0.430 & 0.763 & 0.851 & 0.681 \\
\method{}   & \textbf{0.441} & \textbf{0.782} & \textbf{0.855} & \textbf{0.693} \\
\bottomrule
\end{tabular}
\end{table}

\textbf{Training-bank construction.} We test whether the three-route bank construction of Section~\ref{sec:method-training} teaches the agent to be selective. We build every training core either from source-matched records only (Source-only) or by uniform sampling from the mixed pool (Random), and retrain each variant from SFT onward with all other settings fixed, since the bank shapes both the collected SFT trajectories and the RL rollouts. Besides the original setting, we evaluate on a mixed bank that merges the support banks of MME-R, CharXiv, and MathVision, so that demonstrations from the other two benchmarks act as potential distractors for every query. The two simplified constructions trade accuracy against robustness in opposite ways, as shown in Table~\ref{tab:ablation-bank}. Source-only reaches 0.696 on the own banks, second only to \method{}, but it drops by 2.6 percentage points on the mixed bank, the largest decline among the three variants, because its training banks never required it to reject off-task records. Random degrades least, losing only 0.7 points, yet it starts from the lowest own-bank average of 0.688, as its banks contain few task-matched demonstrations from which to learn how to exploit a good context. The two variants swap ranks between the settings. \method{} achieves the highest score on every benchmark in both settings, keeping 0.693 on the mixed bank after a moderate drop of 1.6 points, so the three-route construction combines the strengths of both simplified variants. The mixed bank also leaves \method{} well ahead of the baselines in our main experiments, even though those baselines retrieve from clean single-benchmark banks. With Qwen3.5-9B, \method{} on the mixed bank still exceeds the strongest baseline, ProMSA, by 3.6 points on MME-R, 4.6 points on CharXiv, and 2.6 points on MathVision, and it surpasses MMICES, ContextNav, and LSD by at least 4.5 points on every benchmark. Thus, its advantage does not depend on a curated bank, and the agent remains effective when relevant demonstrations must be found among distractors.

\textbf{Failure Mode Analysis}
We identified two common failure modes of ActICL based on the evaluation results.
\begin{itemize}
\item Insufficient demonstration coverage. The agent cannot provide complete task guidance when the demonstration bank lacks examples that clarify an uncommon convention or a required combination of operations. It can retain relevant demonstrations of individual steps while leaving the critical connection between those steps unresolved. The answering model must then infer the missing rule from its own knowledge and fails if it applies an inappropriate procedure. Although the agent can reformulate retrieval goals and revise the retained context, it cannot retrieve evidence that the supplied bank does not contain.

\item Errors in reading query-specific visual details.
The answering model produces an incorrect answer when it misreads a decisive detail in the query, even when the retained demonstrations establish the appropriate solution procedure. In dense charts or diagrams, it can confuse a small numerical annotation, an axis unit, or the correspondence between a legend and a visual element. After extracting the wrong value or relation, the model carries that error into an otherwise valid calculation. The answerer must recover these details from the query itself. More demonstrations of the same procedure do not supply the missing query-specific evidence.
\end{itemize}

\section{Prompts and Input Construction}\label{app:prompt}
At the first acquisition step, the trained MLLM agent receives this prompt for a 4-demonstration request under the standard budget.

\begin{ActPromptBox}{Context Acquisition}
\begin{lstlisting}[style=ActCode]
[Acquisition instructions]
You are the context acquisition agent for multimodal in-context learning.
Select complete input-output demonstrations that will help an answering
model solve the query in a separate call. You may reason about how the
query could be solved to identify what guidance is missing, but do not
answer the query.

Identify the task from the query's images and text. Useful demonstrations
clarify the task, solution procedure, output format, or label and symbol
conventions, and each should add guidance beyond the rest of the context.
Visual similarity alone does not establish usefulness; the bank may
contain unrelated tasks, look-alike distractors, and incompatible label
mappings.

At each step, review the query, acquired demonstrations, unread previews,
interaction history, and remaining resources, and decide what is still
missing. Then choose the single action that best addresses it:
- search when no available candidate is likely to supply the missing
  guidance. Match the goal to the need: use alpha when visual appearance
  defines the task, v for visual content you have not yet seen, and ell
  when the task type, question type, or solution procedure is missing.
  Disable channels that would mainly retrieve look-alike but irrelevant
  examples. If a search returns such examples, revise the mode or goal
  instead of repeating it.
- read when a preview suggests that a candidate may supply missing
  guidance. Previews show only a reduced input (downsampled images,
  truncated text) and never the output, so confirm usefulness, including
  the output format and labels, by reading. Skip candidates whose previews
  are clearly irrelevant.
- curate to retain read demonstrations, and to replace retained ones that
  are redundant, misleading, or weaker than an acquired alternative.
  Reading alone does not retain an example; the in_context flag marks
  retained ones. You may keep revising after the requested slots are filled.
- solve when the retained context is admissible and further acquisition
  is unlikely to justify its cost.

A context is admissible if it fits the context-token budget and, when the
requested count is an integer, contains exactly that many demonstrations.
Never exceed the requested count. If strong candidates are scarce, retain
those least likely to mislead. If the count is null, choose how many to
retain, including zero when no candidate helps.

Every action consumes one call. Acquisition tokens include all input and
generated tokens, including images, repeated observations, and your
reasoning, so keep reasoning concise. Rejected calls also consume
resources; read the error and correct the request. Plan with the remaining
resources shown. If no further call fits, the environment commits the
current context only if it is admissible; otherwise acquisition fails.

For each response, enclose your reasoning in <think> and </think>, then
emit exactly one JSON object enclosed in <tool_call> and </tool_call>.
The object must have the keys "name" and "arguments". Use one of the four
tools defined below with its specified arguments. Use valid JSON with
double-quoted keys and strings. Do not emit multiple calls or an answer
to the query.

[Tool definitions]

search(m, alpha, v, ell)
Retrieve up to six previously unseen demonstration-input previews.
Arguments:
- m: "visual", "textual", or "joint".
- alpha: a list of image references; use [] to disable image-to-image
  retrieval. Reference a query image as {"source":"query","image":1},
  or an acquired image as {"source":"acquired","index":1,"image":1}.
  Set the integers to the actual local index and image position. Image
  positions start at 1. Reference only query images or images from
  demonstrations you have already read.
- v: a natural-language description of the visual content you need;
  use "" to disable text-to-image retrieval.
- ell: a natural-language description of the demonstration task or
  solution procedure you need; use "" to disable text-to-text retrieval.
In visual mode, enable alpha, v, or both and leave ell empty. In textual
mode, supply ell and use alpha=[] and v="". In joint mode, enable any
nonempty combination of the three channels. At least one channel must
be active. Describe the desired guidance specifically enough to retrieve
useful examples; do not request a demonstration's hidden target output.
The environment assigns stable local indices to returned candidates.
Earlier unread previews remain available after later searches.

read(index)
Inspect one unread candidate's complete input-output demonstration.
Arguments: "index" must be a positive integer from the unread pool.
The environment moves that candidate from unread previews to acquired
demonstrations. It does not add the demonstration to the retained context.

curate(add, remove)
Revise the retained context using demonstrations already acquired.
Arguments: "add" and "remove" must be lists of local integer indices;
use [] for an empty list. Add only acquired examples not currently
retained. Remove only retained examples. Do not repeat an index or put
it in both lists. The environment checks the resulting context against
the token budget and the fixed-count upper bound, when specified, then
applies both lists together. Removed examples remain acquired, so you
can restore them without another read.

solve()
Commit the current retained context and end acquisition.
Arguments: {}.
The context must fit its token budget and contain exactly the requested
number of demonstrations when the count is specified. Otherwise the
environment rejects the call and acquisition may continue. After a
successful commitment, the answering model receives ICL instructions,
the query, and the retained demonstrations in their first-read order.

After each action, inspect the feedback fields "status", "indices", and
"error". Use only indices currently available in the corresponding pool.
An executed search may return no candidates; an empty result does not
authorize inventing indices or accessing unobserved bank entries.

[Query]
{query_images_and_text}

[Acquired demonstrations: first-read order]
[]

[Unread candidates: first-exposure order]
[]

[Interaction history: chronological order]
[]

[Remaining resources]
calls: 20
acquisition_tokens: 196608
context_tokens: 16384
demonstration_slots: 4

[Requested count]
4
\end{lstlisting}
\end{ActPromptBox}

After context commitment, the trained agent or another MLLM receives this prompt in a fresh answering call, here with two retained demonstrations.

\begin{ActPromptBox}{Answering from the Committed Context}
\begin{lstlisting}[style=ActCode]
[ICL instructions]

Answer the query using its images and text together with the provided
input-output demonstrations. Follow the answer format requested by the
query and output only the final answer.

[Demonstration]
input: {first_retained_demonstration_images_and_text}
output: {first_retained_demonstration_target}

[Demonstration]
input: {second_retained_demonstration_images_and_text}
output: {second_retained_demonstration_target}

[Query]
{query_images_and_text}
\end{lstlisting}
\end{ActPromptBox}

\section{Case Trajectory}\label{app:case}
We illustrate the acquisition under the Standard tier for the chart query of Figure~\ref{teaser}:
``In which year did Online sales grow by the largest percentage over the previous year?'' 
\begin{description}
    \item[Step 1.] The context is empty, so the agent starts from charts like the query. It calls \texttt{search} in visual mode with query image~1 as its reference and leaves $v$ and $\ell$ empty. The environment returns input previews with local indices 1--6.
    \item[Step 2.] The agent calls \texttt{read(1)}, whose preview suggests that it shows which color denotes which series. The environment moves its complete input--output pair into $E_2$, leaving $S_2=\varnothing$.
    \item[Step 3.] The agent calls \texttt{read(2)}, which reads the value of a single bar. The environment adds that demonstration to $E_3$ and removes its preview from the unread pool.
    \item[Step 4.] Series mapping and value reading are both needed, so the agent calls \texttt{curate} with \texttt{add=[1,2]} and \texttt{remove=[]}, retaining both acquired examples in $S_4$.
    \item[Step 5.] Having acquired chart-value reading but nothing on percentage change, the agent calls \texttt{search} in joint mode with $\alpha=[\,]$, $v=$ ``grouped bar charts with color-coded series across years,'' and $\ell=$ ``questions comparing increase (\%) between years.'' The environment returns new candidate indices 7--12 while preserving the earlier unread previews.
    \item[Step 6.] The agent calls \texttt{read(7)}, which reads values and computes percentage growth. The environment now holds acquired indices $[1,2,7]$ in first-read order without changing the retained context.
    \item[Step 7.] Demonstration 2 is subsumed by demonstration 7, so the agent calls \texttt{curate} with \texttt{add=[7]} and \texttt{remove=[2]}. The environment applies the replacement atomically, retaining indices $[1,7]$ while keeping demonstration 2 available in $E_7$. It retains unread indices $[3,4,5,6,8,9,10,11,12]$ and records 13 remaining calls under $H=20$.
    \item[Step 8.] Series mapping and percentage growth are covered, so the agent calls \texttt{solve()} to commit the admissible context. The environment starts a fresh answering call containing the ICL instructions, the complete demonstrations corresponding to indices 1 and 7 in that order, and the query. It omits acquisition indices and history from this answering input.
\end{description}
\end{document}